\pdfoutput=1
\documentclass[11pt]{article}
\usepackage[]{main}

\usepackage{longtable}
\usepackage{array}
\usepackage{rotating}
\usepackage{amsmath}
\usepackage{times}
\usepackage{latexsym}
\usepackage{multirow}
\usepackage{booktabs}
\usepackage{soul}
\usepackage{hyperref}
\usepackage[T1]{fontenc}
\usepackage{todonotes}
\usepackage[utf8]{inputenc}
\usepackage{microtype}
\usepackage{graphicx}
\usepackage{amssymb}
\usepackage{arydshln}

\definecolor{ForestGreen}{RGB}{34,139,34}

\usepackage{array} 
\usepackage{colortbl} 
\usepackage{pdflscape} 

\newcolumntype{Y}{>{\centering\arraybackslash}X}

\usepackage{times}
\usepackage{latexsym}
\usepackage[T1]{fontenc}
\usepackage[utf8]{inputenc}
\usepackage{microtype}
\usepackage{inconsolata}
\usepackage{enumitem}
\usepackage{caption}
\usepackage{subcaption}

\usepackage{xcolor}
\usepackage{soul}

\title{Can LLMs replace Neil deGrasse Tyson?\\Evaluating the Reliability of LLMs as Science Communicators}

\author{$\text{Prasoon Bajpai}^{1}$,
$\text{Niladri Chatterjee}^{1}$, $\text{Subhabrata Dutta}^{1}$, 
$\text{Tanmoy Chakraborty}^{1}$\\
        $^{1} \text{Indian Institute of Technology, Delhi}$\\
        \{prasoonbajpai786, subha009\}@gmail.com, \{niladri, tanchak\}@iitd.ac.in}

\begin{document}
\maketitle

\begin{abstract}
Large Language Models (LLMs) and AI assistants driven by these models are experiencing exponential growth in usage among both expert and amateur users. In this work, we focus on evaluating the reliability of current LLMs as science communicators. Unlike existing benchmarks, our approach emphasizes assessing these models on scientific question-answering tasks that require a nuanced understanding and awareness of answerability. We introduce a novel dataset, {\tt SCiPS-QA}, comprising 742 Yes/No queries embedded in complex scientific concepts, along with a benchmarking suite that evaluates LLMs for correctness and consistency across various criteria. We benchmark three proprietary LLMs from the OpenAI GPT family and 13 open-access LLMs from the Meta Llama-2, Llama-3, and Mistral families. While most open-access models significantly underperform compared to GPT-4 Turbo, our experiments identify Llama-3-70B as a strong competitor, often surpassing GPT-4 Turbo in various evaluation aspects. We also find that even the GPT models exhibit a general incompetence in reliably verifying LLM responses. Moreover, we observe an alarming trend where human evaluators are deceived by incorrect responses from GPT-4 Turbo.

\end{abstract}
\section{Introduction}

The surge of Large language models (LLMs) \cite{brown_few_shot, chowdhery2022palm, chung2022scaling, openai} marks the beginning of an era of rapid development across a variety of natural language tasks. With the introduction of chatbots powered by instruction-tuned LLMs, users across diverse domains are becoming reliant on them in day-to-day activities. The increasing usage of LLM-based AI assistants in academia has triggered intense discussion recently. Multiple reports of inconsistent fragments of text appearing in scientific papers, apparently generated by AI assistants and overlooked due to lack of caution, have surfaced. Recent attempts have been made to outline the usage of AI assistants for literature surveys in research pipelines \cite{bhayana2024chatbots, whitfield2023elicit}

\begin{figure}[t!]
    \centering
    \resizebox{\linewidth}{!}{
    \begin{tabular}{l}
         \includegraphics[width=0.9\textheight,height=1.7\textwidth]{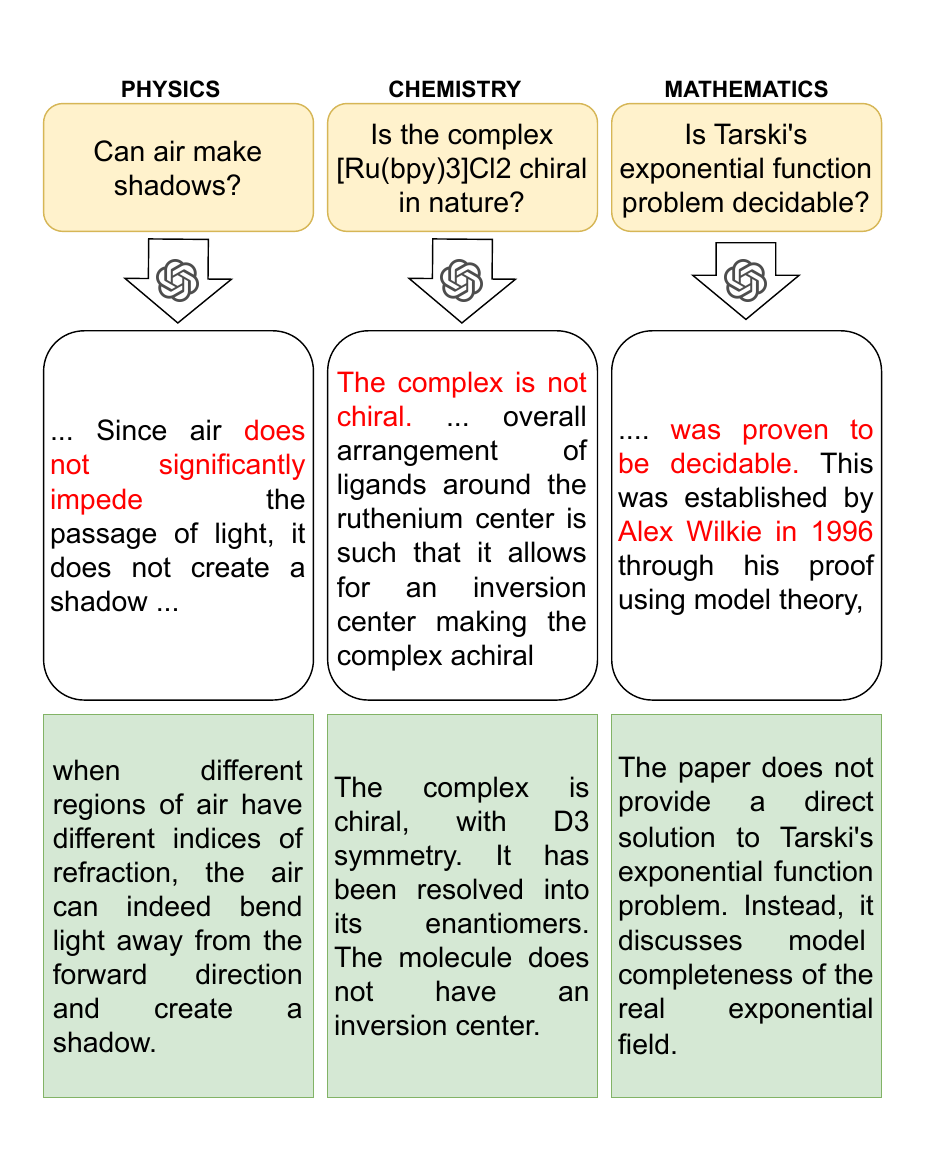}
    \end{tabular}
    }
    \caption{Examples of wrong reasonings given by GPT-4 Turbo to problems in {\tt SCiPS-QA}: (Physics -- Air can cast a shadow under conditions of non-uniform refractive index \cite{physics_example}; Chemistry -- The complex is chiral with D3 symmetry \cite{chemistry_example}; Mathematics -- The paper discusses the model completeness of the real exponential field and its connection to Tarski's problem and the first root conjecture. Tarski's problem is an open problem \cite{mathematics_example}).}
    \label{fig:problem_examples}
    \vspace{-5mm}
\end{figure}

However, there are innate risks associated with LLMs, attributed to overconfident generation and hallucination, that need to be addressed before their large-scale usage as surrogates of human expertise. Particularly in scientific communication in which nuance plays a vital role, LLMs missing out on small details can spread misconceptions~\citep{scientific_miscomm}. Another key challenge lies in the lack of self-awareness of current LLMs and overconfident generation leading to hallucination; given that realizing the lack of knowledge drives the pursuit of scientific exploration, an uncompromising quality of an AI assistant would be to reflect on the lack of knowledge. Existing STEM benchmarks, despite a variety in problem hardness, fail to incorporate these crucial characteristics.

In this work, we seek to close the gaps in evaluating LLMs towards faithful scientific question-answering. Specifically, we seek to address the following research questions:
\begin{itemize}[noitemsep,topsep=0pt]
\item {\bf RQ1:} Can existing LLMs answer scientific reasoning questions successfully and faithfully that require understanding the nuances of scientific knowledge? 
\item {\bf RQ2.} Are LLMs effective at abstaining from assertively answering scientific open problems? 
\item {\bf RQ3.} Can LLMs successfully verify LLM-generated responses? 
\item {\bf RQ4.} Can human evaluators be misled by incorrect yet confident LLM responses to complex scientific questions?
\end{itemize}

To this end, we propose a novel dataset scientific QA dataset, \texttt{SCiPS-QA} (\textbf{S}pecially \textbf{C}halleng{\bf i}ng \textbf{P}roblems in \textbf{S}cience -- {\bf Q}uestion {\bf A}nswering), a collection of $742$ complex boolean scientific problems that require deep knowledge retrieval and extensive reasoning to answer ({\bf Contribution \#1}).\footnote{Please find the code and data at the github repo : \href{https://github.com/Prasoon1207/llm-science-miscommunication}{llm-science-miscommunication}} The problems are chosen from the most niche research areas across different subjects (see Figure~\ref{fig:problem_examples} for sample questions from \texttt{SCiPS-QA} and answers generated by GPT-4 Turbo). \texttt{SCiPS-QA} contains {\em closed} (i.e., the answer exists within the scope of current scientific knowledge) as well as {\em open} problems.  We benchmark a wide variety of proprietary and open-access LLMs from the OpenAI GPT series, Llama-2 and Llama-3 series, and Mistral series on \texttt{SCiPS-QA} using an exhaustive evaluation suit to judge their correctness, faithfulness, and hallucination, in terms of the final boolean answer as well as the reasoning explanation ({\bf Contribution \#2}). We find that while proprietary models like GPT-4 Turbo are generally better than open-access Llama-2, Mistral, or smaller Llama-3 variants, Llama-3-70B models (with or without instruction tuning) come as a strong competitor to GPT-4 Turbo ({\bf Findings \#1}). However, all the experimented LLMs are far from understanding the nuances of scientific rigor, particularly in relation to open problems ({\bf Findings \#2}). We investigate whether proprietary LLMs can successfully verify LLM-generated responses to these complex scientific questions  ({\bf Contribution \#3}), revealing their shortcomings in verifying different aspects of the generated response ({\bf Findings \#3}). Finally, we perform a human evaluation of GPT-4 Turbo generated responses to a subset of questions from \texttt{SCiPS-QA} ({\bf Contribution \#4}). Alarmingly, the persuasive style of generation adopted by GPT-4 Turbo is enough to deceive human evaluators to trust the reasoning, particularly when answers are included in the response ({\bf Findings \#4}).
\section{Related Work}

LLMs have demonstrated various types of reasoning capabilities, including logical, commonsense, mathematical, and temporal reasoning \cite{huang2023reasoning}. In this section, we review relevant work that explores the limitations of LLMs in scientific reasoning.

Several datasets provide comprehensive assessments of LLMs' abilities to solve mathematical problems. GSM8K \cite{cobbe_gsm8k} comprises high-school-level math word problems, while AQuA-RAT \cite{ling-etal-2017-program} includes a collection of algebraic word problems. Dolphin18K \cite{huang-etal-2016-well} features elementary-level problems designed to evaluate basic mathematical reasoning capabilities. The MATH dataset \cite{NEURIPS_DATASETS_AND_BENCHMARKS2021_be83ab3e} presents more challenging problems than those in the aforementioned datasets but focuses on simpler mathematical objects compared to the complex scientific concepts found in \texttt{SCiPS-QA}. Additionally, Ape210K \cite{zhao2020ape210k} offers a broad range of mathematical problems to further test LLMs' problem-solving skills. These datasets collectively highlight the strengths and limitations of LLMs in mathematical reasoning, providing a foundation for understanding their performance in more specialized scientific domains.

ScienceQA \cite{science_qa_lu}, SciQ \cite{science_qa_lu} and MMLU \cite{hendrycks_mmlu} are prominent datasets used to evaluate LLMs' scientific reasoning capabilities. MMLU-Pro \cite{wang_mmlupro} is an improved version of MMLU, offering more challenging problems and greater resistance to prompt variations. ScienceQA is a large-scale multimodal dataset with $21,208$ multiple-choice questions covering diverse science topics. In contrast, SciQ comprises 13.7K multiple-choice science exam questions created through crowdsourcing. We demonstrate that GPT-4 Turbo performs better on these popular STEM datasets than on \texttt{SCiPS-QA}.

\texttt{SCiPS-QA} also focuses on benchmarking answer abstinence in LLMs by including open scientific queries in Physics, Chemistry, and Mathematics. \citealp{dont_hallucinate_feng} explored various answer abstinence methods, evaluating them on MMLU. \citealp{abstention_science_qa_wen} investigated the ability of LLMs to abstain from answering context-dependent science questions when provided with insufficient or incorrect context. They used datasets such as ScienceQA, OpenBookQA, ARC (AI2 Reasoning Challenge) \cite{arc_ai2_clark}, and QASPER \cite{qasper_dasigi} to study LLM abstention behavior. ScienceQA includes school-level and college-level scientific problems requiring relatively simple reasoning capabilities. OpenBookQA features straightforward open-book style general and scientific reasoning problems, and ARC contains grade-school level multiple-choice science questions. In contrast, \texttt{SCiPS-QA} provides a much tougher benchmark for scientific reasoning and explores answer abstinence in Boolean questions.

\section{The \texttt{SCiPS-QA} Dataset}
In this section, we describe the composition of \texttt{SCiPS-QA} and the methodology used to collect the Boolean queries that constitute \texttt{SCiPS-QA}.


\begin{table}[!t]
\centering
\scalebox{1}{
\begin{tabular}{lrrr} 
\hline
\textbf{Subject} & \textbf{Closed} & \textbf{Open} & \textbf{Total}  \\ 
\hline
Physics          & 195                       & 47                      & 242             \\
Chemistry        & 132                       & 0                       & 132             \\
Mathematics      & 140                       & 143                     & 283             \\
Theoretical CS   & 26                        & 22                      & 48              \\
Astronomy        & 15                        & 0                       & 15              \\
Biology          & 1                         & 14                      & 15              \\
Economics        & 1                         & 6                       & 7               \\ 
\hline
\textbf{Total}   & \textbf{510}              & \textbf{232}            & \textbf{742}    \\\hline
\end{tabular}}
\caption{Composition of {\tt SCiPS-QA}.}
\label{tab:dataset_structure}
\vspace{-5mm}
\end{table}
The dataset comprises 742 complex Yes/No problems that require expert-level proficiency in scientific reasoning to answer correctly. We include both open and closed problems across subjects -- Physics, Chemistry, Mathematics, Theoretical Computer Science, Astronomy, Economics, and Biology. Table \ref{tab:dataset_structure} provides the composition of \texttt{SCiPS-QA}, while Figure \ref{fig:domain_problems} shows the subject-level topic decomposition. The difficulty of the problems in the dataset is deliberately kept very high to rigorously test the scientific reasoning and Boolean answering capabilities of state-of-the-art open-source and proprietary LLMs.

\begin{figure}[!t]
    \centering
    \includegraphics[width = \linewidth]{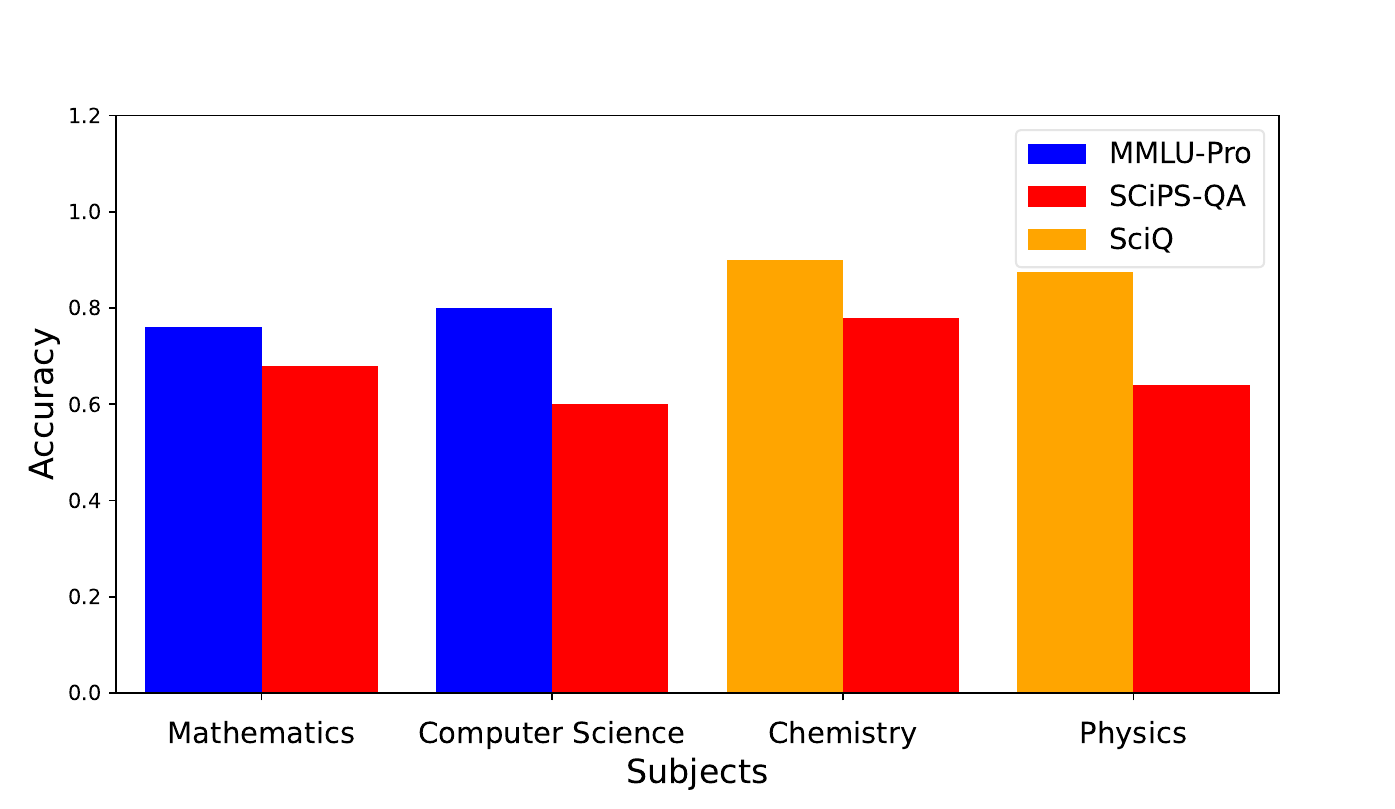}
    \caption{Performance of GPT-4 Turbo on a random subset (of size 40) of MMLU-Pro, SciQ and {\tt SCiPS-QA}. GPT-4 Turbo performs worst on {\tt SCiPS-QA} across all subjects.}
    \label{fig:difficulty_comparison}
    \vspace{-5mm}
\end{figure}

We randomly select 40 problems from each of four different subjects within \texttt{SCiPS-QA} to compare GPT-4 Turbo's performance in answering Boolean scientific queries against those from MMLU-Pro and SciQ.
Additionally, we utilize GPT-3.5 Turbo to paraphrase 40 randomly chosen scientific problems per subject from MMLU-Pro and SciQ into a Yes/No format. Figure \ref{fig:difficulty_comparison} illustrates that GPT-4 Turbo performs the worst on \texttt{SCiPS-QA}, highlighting its higher level of difficulty regarding boolean question answering.\\

\noindent \textbf{Closed questions.} These questions have definitive answers supported by scientific literature. We curate a list of complex topics for each subject in \texttt{SCiPS-QA} manually. For each topic, we utilize the \texttt{wikipedia} API to retrieve its summary. Subsequently, we provide this summary to GPT-4 Turbo, prompting it to generate Yes/No problems along with their corresponding answers. The resulting Boolean questions undergo manual assessment based on two criteria: (1) requiring scientific reasoning for accurate answers, and (2) correctness of the generated answers. The precise prompt used to generate closed questions can be found in Appendix \ref{sec: appendix-closed-questions}.\\

\noindent \textbf{Open questions.} These questions lack a definitive answer in the scientific literature. They are manually selected from \texttt{wikipedia} pages and research blogs. Further details on how open questions were collected can be found in Appendix \ref{sec: appendix-open-questions}.

\section{Experiments}
This section presents the details of the experiments we performed to answer the research questions (RQs) we set to explore.

\subsection{Experimental Setup}

We evaluate a total of 13 open-source models, including those from the Llama-2 family \cite{touvron2023llama}, Llama-3 family, Mistral-7B-Instruct-v0.1 \cite{jiang2023mistral}, Mistral-7B-Instruct-v0.2, and Mistral-8x7B-Instruct-v0.1 \cite{jiang2024mixtral}, on the {\tt SCiPS-QA} dataset using custom-designed evaluation metrics. Additionally, we assess proprietary models such as GPT-4 Turbo (gpt-4-turbo-2024-04-09), GPT-3.5 Turbo (gpt-3.5-turbo-1106), and `text-davinci-003'. For proprietary models, we follow the methodology outlined in \cite{li2024leveraging} to evaluate the reasoning passages generated by these models in response to boolean queries from {\tt SCiPS-QA}. Evaluation criteria include attributes like {\em factuality} and {\em convincingness} (defined in Appendix \ref{sec: definitions}), assessed using GPT-3.5 Turbo and human experts as evaluators. We also evaluate the propensity for {\em hallucination} (see Appendix \ref{sec: definitions})  in these reasoning passages using SelfCheckGPT \cite{manakul2023selfcheckgpt}, which employs a sampling-based approach. Further details on these evaluations are provided in subsequent sections. We collect responses through few-shot prompting. The details about exact prompts can be found in Appendix \ref{sec: appendix-collect-response}. For each model, the responses are collected in two different settings. We call responses collected at temperature 0.0 the `main responses' and those collected at temperature 1.0 the `stochastic responses'.




\subsection{Evaluation Metrics}

Towards a comprehensive evaluation of LLMs, we define the following metrics on the generated responses.

(i) {\bf Main Response Accuracy (MACC).} The accuracy of responses obtained at zero temperature.

(ii) \textbf{Major Stochastic Response Accuracy (MSACC).} We collect the majority response from 10 different stochastic responses with temperature set to 1. We treat invalid responses as incorrect answers.

(iii) {\bf Variation in Stochastic Responses (VSR).} We report the variety in the 10 stochastic responses obtained at temperature 1. We map $ A \rightarrow 1 $, $B \rightarrow 2$, $C \rightarrow 3$ and rest of the invalid responses to 3 and calculate the standard deviation.

(iv) \textbf{Accuracy of Main responses for closed questions (CMACC)} denotes the MACC score on the subset of {\tt SCiPS-QA} containing closed questions.

(v) \textbf{Accuracy of Major Stochastic Responses for Closed Questions (CMSACC)} reflects whether the majority of the LLMs' responses to the closed questions in a unit temperature decoding are correct or not.

(vi) \textbf{Accuracy of Main Responses for Open Questions (OMACC)} is similar to CMACC but evaluated on the open questions instead. In addition to the overall correctness, this metric evaluates whether the model can identify if a question is scientifically unanswerable.

(vii)\textbf{Accuracy of Major Stochastic Responses for Open Questions (OMSACC)} tests the answer abstinence of the LLM in a unit temperature generation regime.

\subsection{Hallucination Quantification}
We employ SelfCheckGPT \cite{manakul2023selfcheckgpt}, a sampling-based methodology that assigns hallucination scores within the range of [0, 1] (0: no hallucination and 1: full hallucination). This scoring is derived by measuring the deviations between the main response and multiple stochastic responses. We take the average of the hallucination scores of sentences in the main response to assign a hallucination score to the entire main response.
We briefly describe each of the SelfCheckGPT variants in this section. More details about SelfCheckGPT and the variants we have implemented can be found in the Appendix \ref{sec: appendix_selfcheckgpt}. 

{\bf SelfCheckGPT with BERTScore.}
For each reasoning sentence in the main response, we calculate the maximum semantic similarity across all sentences in the stochastic response passages. This score indicates the degree of semantic similarity between the main response sentence and various stochastic responses. To quantify hallucination, we derive the complement of this score and assign it as the hallucination score for the main response sentence. Our analysis employs two models, all-MiniLM-L6-v2 and all-mpnet-base-v2, sourced from \texttt{sentence\_transformer} \cite{reimers-2019-sentence-bert}, to generate sentence-level embeddings. This approach ensures mitigation of potential model bias in our results.

{\bf SelfCheckGPT with NLI.}
Natural Language Inference (NLI) assesses whether a hypothesis logically follows from a premise, categorized as entailment, neutral, or contradiction. We compare each sentence of a main response reasoning passage as a hypothesis against each of the corresponding stochastic response reasoning passages as the premise. The logits associated with classes `contradiction' and `entailment' are considered and a score is assigned to the main response sentence, which is a proxy for the probability score of it being in `contradiction' to the stochastic response reasoning passages. We use DeBERTa-v3-base \cite{deberta} fine-tuned on MNLI \cite{williams-etal-2018-broad} for collecting the logits associated with `contradiction' and `entailement' classes.

{\bf SelfCheckGPT with Prompt.}
We employ an external LLM evaluator to determine if each sentence in a main response reasoning passage is supported by corresponding stochastic response reasoning passages. Specifically, we utilize GPT-3.5 Turbo as the external LLM; the exact prompt used can be found in Appendix \ref{sec: appendix-selfcheckgpt-with-prompt}. The responses (Yes, No, NA) are mapped to hallucination scores (Yes $\rightarrow$ 0, No $\rightarrow$ 1, NA $\rightarrow$ 0.5). The average of the GPT-3.5 Turbo response scores is calculated and assigned as the hallucination score for the corresponding sentence in the main response.

\subsection{NLG Evaluation of Reasoning Passages}
We validate the main response reasoning passages generated by the models -- GPT-4 Turbo, GPT-3.5 Turbo, and text-davinci-003 using GPT-3.5 Turbo as the verification model. Additionally, we verify responses from GPT-4 Turbo using GPT-4 Turbo itself as the verification model. Verification attributes are scored on a linear scale using prompt outputs in a zero-shot setting. All relevant prompt details can be found in Appendix \ref{sec: appendix-nlg-eval-prompt}.

{\bf Convince-factor.} Responses that are highly convincing but rely on incorrect information are considered `hallucinations' \cite{hallucination_survey}. We assign a {\em convincingness score} on a linear scale ranging from 1 to 5. This verification attribute is reported for main response reasoning passages using two different prompt settings: one where model answers are included in the prompt given to the evaluator models (denoted as \texttt{convince-factor-with-answer}), and another where model answers are absent (denoted as \texttt{convince-factor-without-answer}).

{\bf Fact-check.} We assign scores on a linear scale (ranging from 1 to 5) to main response reasoning passages based on their factual accuracy. Our aim is to investigate whether evaluator LLMs can differentiate between incorrect reasoning passages and correct ones based on the factual correctness of responses.


{\bf Information Mismatch.} We compare each main response reasoning passage with all ten different stochastic response reasoning passages $S_k$ for the amount of information mismatch between them, which is scored on a linear scale ranging from 1 to 5. We assign the mean of such scores across stochastic responses to the main response reasoning passage. 

\subsection{Human Evaluations}
We randomly select 30 combinations of query and main response reasoning passages (from GPT-4 Turbo) for each subject -- Physics, Chemistry, Mathematics, and Computer Science. For each subject, we employed two human evaluators.
All human evaluators had at least a graduate degree in their respective subjects; they were male and aged between 20-25.

Human evaluators were tasked with assigning a `convince-factor' score to the main response reasoning passages, following the same evaluation setup used with LLMs as the evaluators. We divide human evaluators into two groups: one group sees both the model answer and reasoning, while the other group only views the reasoning itself. Both groups receive identical queries for evaluation. 
\begin{table*}[!t]
\begin{center}
\resizebox{\textwidth}{!}{%
\begin{tabular}{l|cccccccc}
\hline
{\bf LLMs} & \textbf{MACC} (↑) & \textbf{MSACC} (↑) & \textbf{VSR} (↓) & \textbf{CMACC} (↑)& \textbf{CMSACC} (↑) & \textbf{OMACC} (↑) & \textbf{OMSACC} (↑) \\ 
\hline
meta-llama-2-7B & $0.021$ & $0.108$ & $0.922$ & $0.031$ & $0.157$ & $0.000$ & $0.000$ \\
meta-llama-2-7B-chat & $0.321$ & $0.272$ & $1.069$  & $0.284$ & $0.255$ & $0.400$ & $0.310$ \\
meta-llama-2-13B & $0.327$ & $0.361$ & $0.826$ & $0.476$ & $0.523$ & $0.000$ & $0.004$\\
meta-llama-2-13B-chat & $0.341$ & $0.356$ & $0.636$  & $0.484$ & $0.500$ & $0.026$ & $0.039$ \\
meta-llama-2-70B & $0.532$ & $0.274$ & $1.097$ & $0.498$ & $0.292$ & $0.608$ & $0.232$ \\
meta-llama-2-70B-chat & $0.423$ & $0.426$ & $0.689$ & $0.616$ & $0.620$ & $0.000$ & $0.000$ \\ 
\hline
meta-llama-3-8B & $0.120$ & $0.010$ & $1.014$ & $0.174$ & $0.139$ & $0.000$ & $0.004$ \\
meta-llama-3-8B-instruct & $0.444$ & $0.437$ & $0.550$ & $0.645$ & $0.635$ & $0.004$ & $0.000$ \\
meta-llama-3-70B & $\textbf{0.693}$ & $0.605$ & $0.964$  & $0.743$ & $0.659$ & $\textbf{0.582}$ & $\textbf{0.487}$ \\
meta-llama-3-70B-instruct & $0.628$ & $\underline{0.623}$ & $0.295$ & $\textbf{0.780}$ & $\textbf{0.784}$ & $0.293$ & $0.267$ \\ 
\hline
Mistral-7B-Instruct-v0.1 & $0.113$ & $0.311$ & $0.660$ & $0.165$ & $0.453$ & $0.000$ & $0.000$ \\
Mistral-7B-Instruct-v0.2 & $0.496$ & $0.488$ & $0.474$  & $0.582$ & $0.574$ & $0.306$ & $0.297$ \\
Mixtral-8x7B-Instruct-v0.1 & $0.591$ & $0.596$ & $0.555$ & $0.678$ & $0.682$ & $0.401$ & $0.405$ \\ 
\hline
text-davinci-003 & $0.548$ & $0.554$ & $\underline{0.229}$ & $0.723$ & $0.717$ & $0.187$ & $0.216$ \\
GPT-3.5 Turbo & $0.576$ & $0.597$ & $0.337$ & $0.691$ & $0.711$ & $0.340$ & $0.361$ \\
GPT-4 Turbo & $\underline{0.646}$ & $\textbf{0.651}$ & $\textbf{0.193}$ & $\underline{0.750}$ & $\underline{0.754}$ & $\underline{0.432}$ & $\underline{0.436}$\\
\hline
\end{tabular}
}
\end{center}
\caption{Comparative evaluation of state-of-the-art open-source and proprietary LLMs across multiple evaluation metrics. The symbol ↑ (↓) indicates the higher (lower) value is better. We \textbf{bold} the best and \underline{underline} second-ranked score for each metric.}
\label{tab:evaluation_table_main}
\vspace{-5mm}
\end{table*}
\section{Results}
In this section, we look at the various quantitative results summarized in Table \ref{tab:evaluation_table_main}.

\subsection{\texttt{SCiPS-QA} Benchmark}
We observe that among both open-source and proprietary models, the Llama-2 family consistently performs the poorest across nearly all metrics. The GPT series of models show competitive performance, closely rivaling the higher-scale models within the Llama-3 family, which rank highest among the open-source models tested.

\par
\textbf{MACC}: Llama-3-70B achieves the highest score in the MACC metric at $0.693$, closely followed by GPT-4 Turbo with a score of $0.646$. Notably, among the Llama-2 and Llama-3 families, `chat' models perform equivalently to their non-instruction fine-tuned counterparts, except for the lower scale members: Llama-2-7B and Llama-3-8B, where the instruction fine-tuned variants show score increases of $0.3$ and $0.324$, respectively. Mixtral-8x7B-Instruct-v0.1 significantly outperforms both Mistral-7B-Instruct-v0.1 and Mistral-7B-Instruct-v0.2. All three GPT models perform strongly, with GPT-4 Turbo achieving the highest score of $0.646$ in the MACC metric.

\par
\textbf{MSACC}: GPT-4 Turbo outperforms all other models with a score of $0.651$, closely followed by Llama-3-70B-instruct, which achieves a score of $0.623$. In contrast to the MACC metric, where instruction fine-tuned models from the Llama-2 and Llama-3 families often performed equivalent to their non-instruction fine-tuned counterparts, here we observe that the instruction fine-tuned models outperform their counterparts.

\par
\textbf{VSR}: The Llama-2 family performs the worst in terms of VSR score indicating their limited capability to produce consistent results. In contrast, the GPT models exhibit high consistency, with GPT-4 Turbo reporting the lowest VSR score of $0.193$ among all models. The Llama-3 family demonstrates better consistency compared to the Llama-2 family, while Mistral models also perform well but not as strongly as the top performers among open-source models. Among them, Llama-3-70B-instruct stands out with a VSR score of $0.295$.

\par
\textbf{CMACC, CMSACC}: Llama-3-70B-instruct outperforms all models in CMACC and CMSACC metrics achieving a score of 0.780 and 0.784 respectively. GPT models also perform well in handling closed domain scientific queries with GPT-4 Turbo being the best among them, achieving a score of 0.75 and 0.754.

\par
\textbf{OMACC, OMSACC}: One of the major findings is that most of the open source and proprietary LLMs are really bad at accepting that they do not know the answers to open scientific queries in \texttt{SCiPS-QA}. This is evident from their low OMACC and OMSACC scores across the board. Llama-3-70B stands out as the top performer in terms of answer abstention for open scientific queries, achieving the highest OMACC ($0.582$) and OMSACC ($0.487$) scores. In contrast, Llama-2 models struggle significantly in handling open queries, while Mistral-7B models and Mixtral-8x-7B-Instruct-v0.1 perform reasonably well among open models. The GPT models demonstrate strong performance in responding to open scientific queries, with GPT-4 Turbo achieving the highest scores of $0.432$ and $0.436$ in OMACC and OMSACC metrics, respectively.
Note that models also produced invalid responses to prompts. Small models -- Llama-2-7B and Mistral-7B-Instruct-v0.1, produce a much larger fraction of invalid responses as compared to other open-source models. Proprietary models produce almost negligible invalid responses with GPT-4 Turbo reporting no invalid main response. More details can be found in Appendix \ref{sec:appendix-perc-empty}.

\subsection{Hallucination Quantification}
\begin{figure*}[!t]
\includegraphics[width=\textwidth]{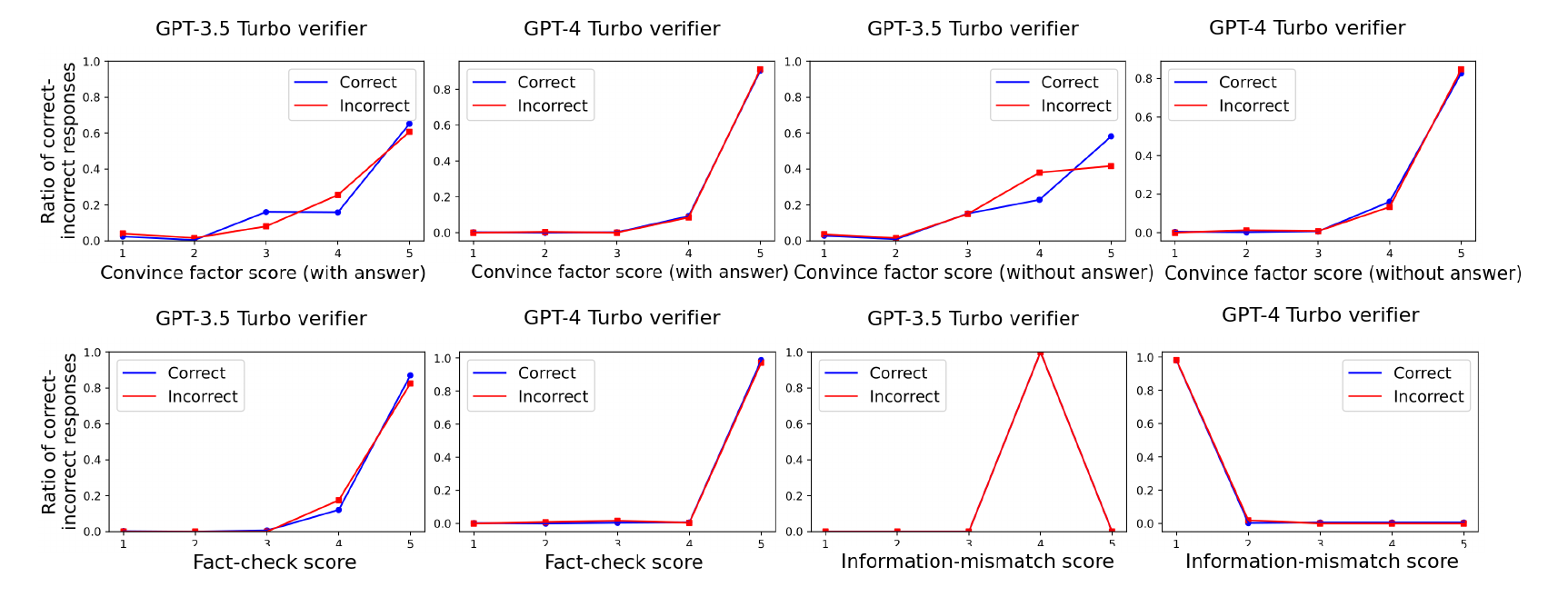}
    \caption{Verification of the reasoning passages generated by GPT-4 Turbo across convincingness (with and without answer), factuality, and information mismatch; we use both GPT-4 Turbo and GPT-3.5 Turbo as verifier models. The fraction of correct (incorrect) responses at each score level is shown in blue (red). An ideal verifier should provide all the incorrect responses with the lowest score (1) and all the correct responses with the highest score (5). However, no verifier model in our experiments could demarcate between the correct and incorrect responses.}
    \label{fig:enter-label}
    \vspace{-5mm}
\end{figure*}
Our investigation using SelfCheckGPT fails to yield conclusive evidence of hallucination in the proprietary GPT models despite their high rate of mistakes. When employing the BERTScore variant, we observe normal distribution in the frequency distribution histograms (Figure \ref{fig:result_bert_score}) for all three GPT models on \texttt{SCiPS-QA}. Interestingly, GPT-3.5 Turbo achieves the lowest mean hallucination score, followed by GPT-4 Turbo, while text-davinci-003 performed the poorest.

Assuming normal distribution and independence of score samples, we conduct Welch's t-tests to assess the statistical significance of mean differences between the proprietary models. Our findings indicate that we fail to reject the hypothesis of no difference in means between all pairs of proprietary models being tested with a 95\% level of confidence. Further details of these tests are available in Appendix \ref{sec: appendix-result-hall-quant}.

Figure \ref{fig:selfcheckgpt_prompt_result} summarizes results for SelfCheckGPT with Prompt. GPT-3.5 Turbo shows a higher count of low hallucination scores given to queries in \texttt{SCiPS-QA} than GPT-4 Turbo.
\subsection{NLG Evaluation of Reasoning Passages}

We assess the main response reasoning passages from all three proprietary models using GPT-3.5 Turbo as the verifier. Table \ref{tab:nlg_evaluation_result} shows the convincingness (with and without answer), factuality and information-mismatch scores for all three models using GPT-3.5 Turbo as verifier. We use both GPT-4 Turbo and GPT-3.5 Turbo for verifying reasoning passages obtained from GPT-4 Turbo. Figure \ref{fig:enter-label} shows the verification results for the GPT-4 Turbo’s reasoning passages for the two verifier models.

\subsubsection{Convince-factor}
GPT-3.5 Turbo consistently assigns high scores to the main response reasoning passages from all three models (see Table \ref{tab:nlg_evaluation_result}). It rates both correct and incorrect reasoning responses highly across all models. Surprisingly, even when evaluating reasoning passages from GPT-4 Turbo, it itself struggles to distinguish between correct and incorrect responses.
Interestingly, as depicted in Figure~\ref{fig:enter-label}, GPT-4 Turbo assigns a higher fraction of reasoning passages (both correct and incorrect) a perfect score of 5 in convincingness (with and without answer) compared to GPT-3.5 Turbo. This suggests that GPT-4 Turbo performs worse than GPT-3.5 Turbo in terms of verifying its responses based on convincingness (with and without answer).

\subsubsection{Fact-check}
GPT-3.5 Turbo assigns high scores to the reasoning passages from all three models (see Table \ref{tab:nlg_evaluation_result} in Appendix), often rating a majority of incorrect responses a perfect 5 in factuality verification. GPT-4 Turbo performs even worse in verifying its own reasoning passages (see Figure \ref{fig:enter-label}), assigning a higher fraction of incorrect reasoning passages a perfect 5 score compared to GPT-3.5 Turbo. This indicates that GPT-4 Turbo struggles more than GPT-3.5 Turbo in distinguishing between correct and incorrect reasoning passages, even when evaluating its own responses.

\subsubsection{Information Mismatch}

We observe that GPT-3.5 Turbo assigns relatively high \texttt{information-mismatch} scores to main response reasoning passages from all three proprietary models. Table \ref{tab:nlg_evaluation_result} shows that among the three models being tested, GPT-3.5 Turbo gives a lesser information-mismatch score to its own reasoning passages than it does to the other two models.

From Figure \ref{fig:enter-label}, we observe that GPT-4 Turbo provides a very low score for its own reasoning passages in terms of \texttt{information-mismatch} score. These patterns are agnostic to the correctness of reasoning passages, suggesting that both verifier models are not able to differentiate between correct and incorrect passages using \texttt{information-mismatch} scores. Consistent with our observations across various verification attributes, GPT-4 Turbo performs worse than GPT-3.5 Turbo, consistently assigning lower scores (often 1) to most reasoning passages, irrespective of their correctness.

\subsection{Human Evaluations}
\begin{figure}[!t]
\includegraphics[width = \linewidth]{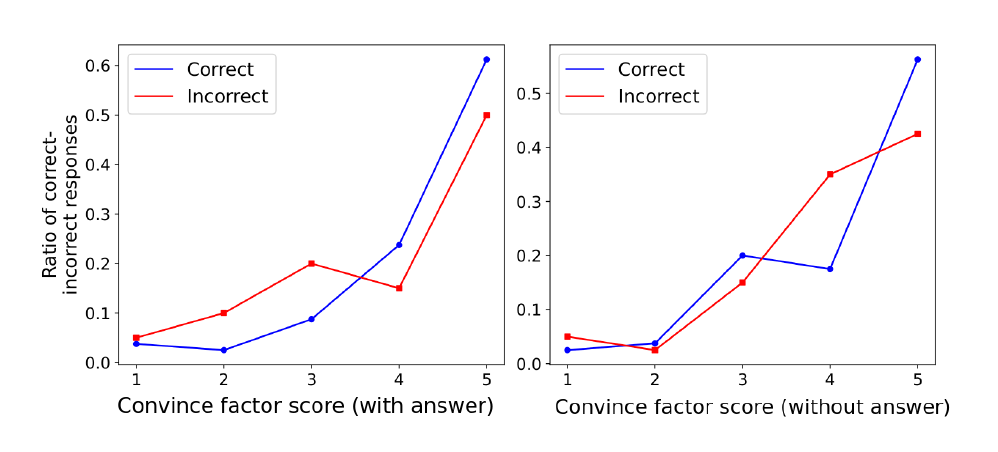}
\caption{Distribution of correct (in blue) and incorrect (in red) responses generated by GPT-4 Turbo against convince factor scores provided by human evaluators. Incorrect LLM reasoning can deceive humans as convincing with or without the answer shown to them. However, humans provide better judgement with the answer.}
\label{fig:result_human_evaluation}
\vspace{-5mm}
\end{figure}

Human evaluators typically fare better than LLM evaluators. As we can see in Figure~\ref{fig:result_human_evaluation}, correct responses are consistently given better scores than incorrect ones. However, a considerable fraction of incorrect responses can still deceive human judgment into getting scores greater than 3. 

Notably, human evaluators tend to judge incorrect responses better when the generated answer is attached. This can be possibly related to cases where the LLM infers incorrect answers even after providing correct reasoning context. Furthermore, correct responses are typically distributed towards the highest convince factor (i.e., 5); although, without the answer provided, some correct responses are given scores as low as 3.
Interestingly, the scoring distribution provided by human evaluators is much closer to that provided by GPT-3.5 Turbo as verifier instead of GPT-4 Turbo.

\section{Discussion and Conclusion}

Our experiments on \texttt{SCiPS-QA} with a diverse array of LLMs using a comprehensive evaluation strategy reveal several key insights. Firstly, existing LLMs, whether open-access or proprietary, demonstrate a limited understanding of scientific methodologies required to serve as reliable assistants. While the parameter scaling law holds within each LLM family, models of similar size across different families are not directly comparable. For instance, Meta Llama-3 70B models emerge as formidable competitors to much larger GPT models, frequently outperforming GPT-4 Turbo in our evaluations. This reiterates earlier findings that parameter scaling alone does not reflect the capabilities of LLMs and current models, along with their training methodology, are underperforming their `true' potential~\cite{hoffmann2022training}.

Echoing \citet{huang2024large}'s findings, we observe that powerful LLMs such as GPT-4 Turbo and GPT-3.5 Turbo struggle to reliably verify their responses. Hallucination detection techniques like SelfCheckGPT also prove ineffective in detecting incorrect reasoning posed by strong LLMs like GPT-4 Turbo in complex questions within \texttt{SCiPS-QA}. In fact, we notice a counterintuitive trend where GPT-3.5 Turbo assigns lower scores to incorrect responses compared to the stronger GPT-4 Turbo. 

However, the most concerning finding of this paper revolves around how human evaluators perceive LLM-generated scientific reasoning. When tasked with evaluating the convincingness of reasoning explanations generated by GPT-4 Turbo, human evaluators tend to assign higher ratings to a significant majority of incorrect answers. This aligns with the concern raised by \citet{scientific_miscomm} that current LLM-based AI assistants have the potential to propagate widespread scientific misunderstandings if left unchecked.

{\bf Implications for future research.} We hope that our proposed dataset, \texttt{SCiPS-QA}, along with the evaluation suit we design in this work, will serve as a valuable benchmark for future LLM research. Given the growing popularity of generalist as well as domain-specific AI assistants, we envision a positive future focus in building reliable scientific assistants. Finally, our findings with human evaluation calls upon further focus in trustworthy AI research.

\newpage
\section{Limitations}
Boolean format of scientific questions has been adopted in \texttt{SCiPS-QA}. Having a long-text reasoning evaluation while maintaining the complexity of scientific objects should provide a stronger test for evaluating scientific communication. For this, \texttt{SCiPS-QA} needs to be augmented with golden reasoning passages provided by human experts. There is also a need to add more diverse topics to the \texttt{SCiPS-QA}, particularly in Physics, which is dominated by Quantum Mechanics (Appendix \ref{fig:domain_problems}). There is also an issue of some queries in \texttt{SCiPS-QA} lying outside the knowledge cutoff of some models, making it difficult to accurately assess their reasoning capabilities. Human evaluations may be slightly limited because they do not include highly experienced evaluators in the respective subjects. The testing of reasoning passages from open-source models has also not been done as part of our analysis. 
\section{Ethical Considerations}

The participants in human evaluation were not coerced into participating and were given clear and comprehensive information about the research before they provided informed consent. The identities of the human evaluators have been protected by ensuring their responses cannot be linked back to the specific individuals. The research results are communicated honestly and credibly and transparency has been maintained throughout the research process.

\bibliographystyle{acl_natbib}
\bibliography{anthology}
\appendix
\newpage
\begin{figure*}[hbt!]
\centering
    \begin{tabular}{cc}
        \includegraphics[width = 0.5\linewidth]{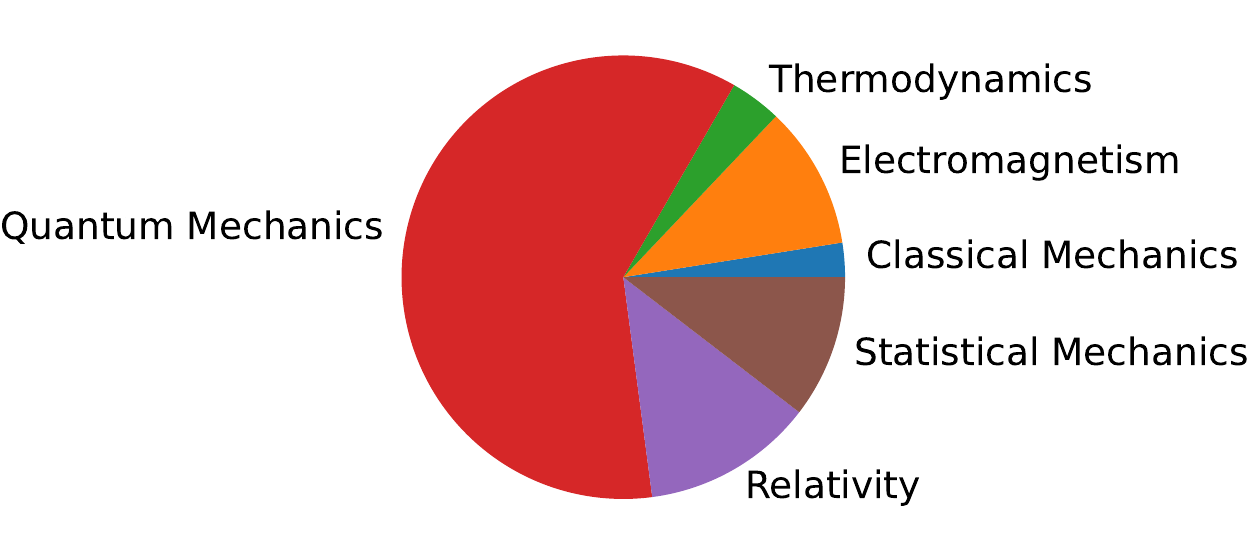}&
        \includegraphics[width = 0.5\linewidth]{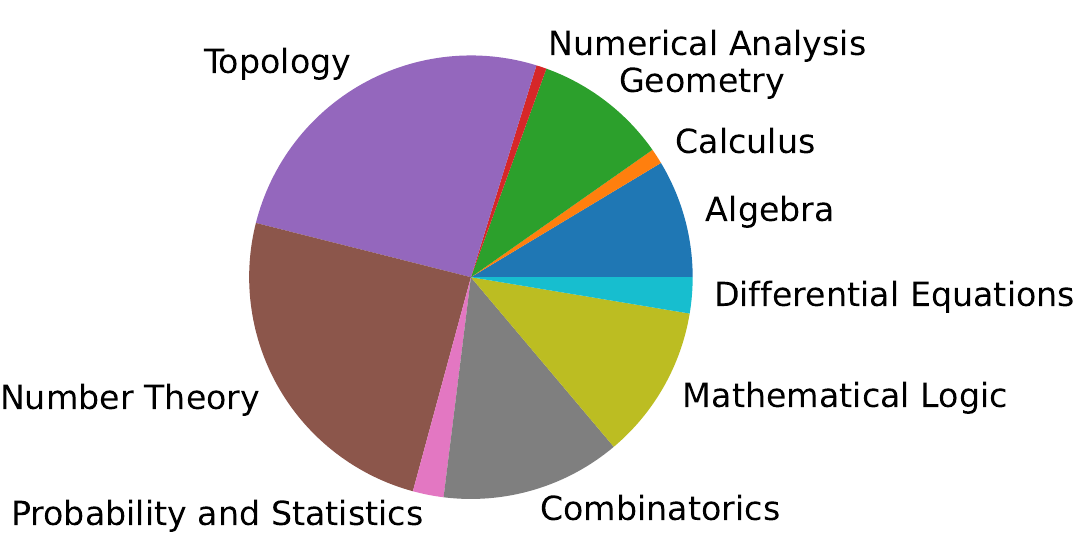}
    \end{tabular} 
    \begin{tabular}{cc}
         \hspace{4cm} &
         \includegraphics[width = 0.7\linewidth]{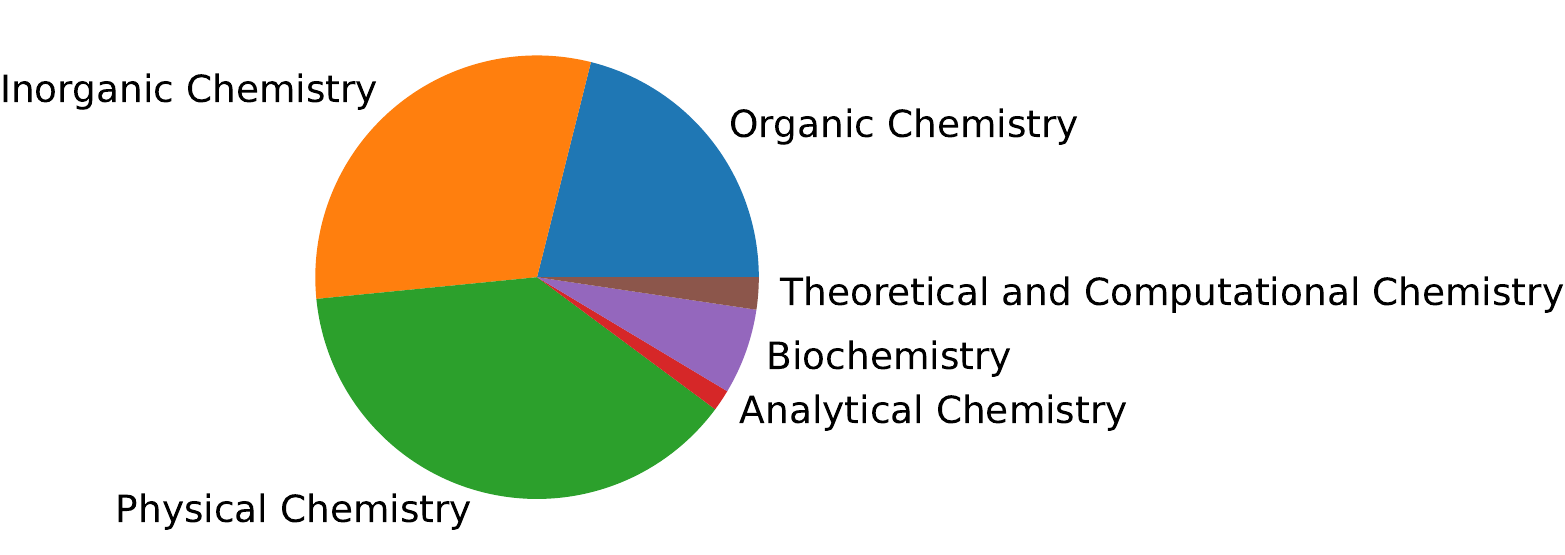}
    \end{tabular}
\caption{Topic decompostion for subjects : Physics (top-left), Chemistry (middle) \& Mathematics (top-right) in \texttt{SCiPS-QA}}
\label{fig:domain_problems}
\end{figure*}




\section{Definitions}
\label{sec: definitions}
\noindent \textbf{Hallucination}:  The generated content that is
nonsensical or unfaithful to the provided source input \cite{filippova2020controlled, maynez2020faithfulness}, where the source input changes as the task. We take the world knowledge as the source input in our case.\\
\noindent \textbf{Factuality}: Factuality refers to the property of quality of being actual or based on fact \cite{dong2020multi}. In our work, we take "facts" as the world knowledge.\\
\noindent \textbf{Convincingness}: Convincingness refers to the ability of a model to effectively influence the audience through language \cite{habernal-gurevych-2016-makes}.

\section{Collection of Open Questions}
\label{sec: appendix-open-questions}
We collect open-questions from \href{https://en.wikipedia.org/wiki/Lists_of_unsolved_problems}{List of unsolved problems} article on wikipedia for all subjects. We also referred to the page \href{https://a3nm.net/work/research/questions/}{List of open questions in theoretical computer science} by \href{https://a3nm.net/}{Antoine Amarilli}. We use GPT-3.5 Turbo to parse some of the entries on these web pages into a question format.  

\section{SelfCheckGPT}
\label{sec: appendix_selfcheckgpt}
\subsection{Notation}
We obtain two types of responses from proprietary models for quantifying hallucination. Let $M$, calling it the ‘main response’, denote the reasoning passage obtained at temperature 0.0. We sample N = 10 different stochastic responses:$\{S_{1}, S_{2}, …S_{N}\}$, each at temperature 1.0 using the same prompt structure, aiming to measure commonalities between the stochastic responses and the main response. We use SelfGPTCheck to assign a hallucination score to ith sentence of the main response $M_{i}$ : $H(M_{i}) -> [0.0, 1.0]$, with 0.0 score given to such sentences that are completely faithful to source input and 1.0 if they are fully hallucinated. The following subsections describe the variants of SelfCheckGPT briefly that we have used in this paper.

\subsection{SelfCheckGPT with BERTScore}
\label{sec:experiment-selfcheckgpt-bert}
Let $M_{i}$ and $S_{j}^{k}$ denote the $i$-th sentence of the main response and the $j$-th sentence of the $k$-th stochastic response. Note all these responses are reasoning passages that are provided by the proprietary models tested. We assign a hallucination score to $M_{i}$ depending on the BERTScore between $M_{i}$ and $S_{j}^{k}$ as follows:\\
\begin{equation}
     H(M_{i}) = 1 - \frac{1}{N} \sum_{k=1}^{N} \max_k B(M_i, S_j^k) 
\end{equation}
where $B(., .)$ is the dot score of sentence embeddings generated using model $B$. This way $M_{i}$ shall be assigned a higher score if it is semantically less similar (according to BERTScore) to most of the sentences in different stochastic responses. However, if a sentence in the main response is semantically similar (or appears in) to sentences in different stochastic responses, then it will be assigned a lower hallucination score. We take the mean of the hallucination scores of each sentence of the main response to assign it a hallucination score.

We report results using two different models : $B \in $ \{\texttt{all-MiniLM-L6-v2}, \texttt{all-mpnet-base-v2}\} from \texttt{sentence$\_$transformer} \cite{reimers-2019-sentence-bert} for generating sentence-level embeddings for eliminating any possible model bias.

\subsection{SelfCheckGPT with NLI}

The input for NLI classifiers is typically the premise concatenated to the hypothesis, which for our methodology is the sampled passage $S_k$ concatenated to the sentence to be assessed $M_i$. Only the logits associated with the ‘entailment’ and ‘contradiction’ classes are considered, We use \texttt{DeBERTa-v3-base} fine-tuned on MNLI for collecting the logits associated with 'contradict' class.

SelfGPTCheck with NLI uses stochastic response $S_k$ as the premise concatenated to the main response sentence $M_i$ to be assessed. The logits associated with token ‘contradict’ are used to assign a score. 
\begin{equation}
P(\text{contradict} | M_i, S_k) = \frac{\exp(z_c)}{\exp(z_e) + \exp(z_c)} 
\end{equation}
where $z_e$ and $z_c$ are the logits of the ‘entailment’
and ‘contradiction’ classes, respectively. A higher probability denotes that the concerned main response sentence disagrees with the stochastic sample and hence, should be assigned a higher hallucination score, which is defined as,

\begin{equation}
    H(M_i) = \frac{1}{N} \sum_{k=1}^{N} P(\text{contradict}|M_i, S_k)
\end{equation}
We take the average of the hallucination scores of sentences in the main response to assign a hallucination score to the entire main response $M$.

\subsection{SelfCheckGPT with Prompt}
We prompt GPT-3.5 Turbo to assess if the $i$-th sentence of the main response is supported by the $k$-th stochastic response, $S_{k}$. The exact prompt can be found in the appendix \ref{sec: appendix-selfcheckgpt-with-prompt}.

The output from prompting when comparing the $i$-th sentence against sample $S_k$ is converted to score $x_i^k$ through the mapping {Yes: 0.0, No: 1.0, N/A: 0.5}. The final inconsistency score is then
calculated as:
\begin{equation}
    H(M_i) = \frac{1}{N}\sum_{k=1}^{N}(x_i^k)
\end{equation}

Note, for all these variants, we report the results at only such data-points of \texttt{SCiPS-QA} where all 10 stochastic reponses are non-empty and valid. A stochastic response is considered invalid if it cannot be parsed into the boolean answer and the corresponding reasoning passage.

\section{Prompts}
\begin{table*}[!ht]
\begin{center}
\small 
\setlength{\tabcolsep}{2pt} 
\renewcommand{\arraystretch}{1.0} 
\resizebox{\textwidth}{!}{%
\begin{tabular}{p{1.2in}@{\hspace{4em}}p{5.2in}}
\toprule
\textbf{\texttt{Collecting closed-questions from \texttt{wikipedia} passage\dag}} &
  \begin{tabular}[l]{@{}p{5.2in}@{}}
  \texttt{You are an AI assistant to create extremely challenging Yes/No problems , from the provided passage.}\\
  \texttt{<PASSAGE>}\\
  \texttt{Generate your response strictly in the following JSON format.}\\
  \texttt{Create questions which specifically require reasoning to answer.}\\ 
  \texttt{\{\newline
    \hspace*{1cm}"questions":\newline
    \hspace*{1cm}[\newline
        \hspace*{1cm}\{\newline
            \hspace*{2cm}"question" : Can a regular expression represent all possible languages over an alphabet?,\newline
            \hspace*{2cm}"answer" : No\newline
        \hspace*{1cm}\},\newline
        \hspace*{1cm}\{\newline
            \hspace*{2cm}"question" : Can every problem in the complexity class EXP be solved by a deterministic Turing machine in exponential time?,\newline
            \hspace*{2cm}"answer" : Yes\newline
        \hspace*{1cm}\},\newline
        \hspace*{1cm}\{\newline
            \hspace*{2cm}"question" : Is it possible to efficiently approximate the solution to an NP-hard optimization problem within a constant factor?,\newline
            \hspace*{2cm}"answer" : No\newline
        \hspace*{1cm}\}\newline
    \hspace*{1cm}]\newline
  \}\newline}\\
  \texttt{Create extremely challenging Yes/No questions, requiring reasoning to answer, from the passage provided below. Remember following points -}\\
  \texttt{1. Use the JSON format specified above.\newline
    2. Create questions that specifically require some reasoning for their resolution.}
  \end{tabular} \\
\midrule
\textbf{\texttt{Collecting responses (open-source models) \ddag}} &
  \begin{tabular}[c]{@{}p{5.2in}@{}}
    \texttt{You are an AI assistant being used for research purposes.\newline
    Answer the following Yes/No multi-choice questions with following options.\newline}
    \texttt{A - Yes\newline
    B - No\newline
    C - I do not know\newline}
    \texttt{
    Question: Is the nature of spacetime at the Planck scale continuous or discrete?\newline
    Answer: C\newline
    Question: Is the exact value of mean lifetime of a neutron known?\newline
    Answer: B\newline
    Question: Does the set of natural numbers have the same cardinality as set of integers?\newline
    Answer: A\newline
    Question: Is the Universe infinite in spatial extent?\newline
    Answer: C}
  \end{tabular} \\
\midrule
\textbf{\texttt{Collecting responses (proprietary models)\ddag}} &
  \begin{tabular}[c]{@{}p{5.2in}@{}}
    \texttt{You are an AI assistant being used for research purposes.\newline
    Answer the following Yes/No multi-choice questions with following options.\newline}
    \texttt{A - Yes\newline
    B - No\newline
    C - I do not know\newline}
    \texttt{
    Question: Is the nature of spacetime at the Planck scale continuous or discrete?\newline
    Reason: The nature of spacetime at the Planck scale is currently not definitively established, and it remains an open question in theoretical physics. However, many theories, including some formulations of quantum gravity, suggest that spacetime may exhibit discrete or quantized characteristics at the Planck scale. This is based on the idea that classical notions of smooth, continuous spacetime may break down at extremely small scales. The uncertainty principle from quantum mechanics also contributes to this speculation. Research in areas such as loop quantum gravity and string theory explores these possibilities, but a conclusive answer is yet to be determined through experimental verification.\newline
    Answer: C\newline
    Question: Is the exact value of mean lifetime of a neutron known?\newline
    Reason:  ... \newline
    Answer: B\newline
    Question: Does the set of natural numbers have the same cardinality as set of integers?\newline
    Reason:  ... \newline
    Answer: A\newline
    Question: Is the Universe infinite in spatial extent?\newline
    Reason: ...\newline
    Answer: C}
  \end{tabular} \\
\bottomrule
\end{tabular}%
}
\end{center}
\caption{Prompts used for collecting closed-questions from Wikipedia pages and collecting responses from open-source models and proprietary models. For 'Collecting responses (proprietary models)', we do not write 'Reason' for all exemplars due to lack of space.\\
\dag - Prompts made to GPT-4 Turbo\\
\ddag - Prompts made to GPT-3.5 Turbo}
\label{tab:prompting_templates_1}
\end{table*}

\begin{table*}[ht]
\begin{center}
\small 
\setlength{\tabcolsep}{2pt} 
\renewcommand{\arraystretch}{1.0} 
\resizebox{\textwidth}{!}{%
\begin{tabular}{p{1.2in}@{\hspace{4em}}p{5.2in}}
\toprule
\textbf{\texttt{SelfCheckGPT with Prompt}} &
  \begin{tabular}[c]{@{}p{5.2in}@{}}
    \texttt{Context: <CONTEXT>\newline
    Sentence: <SENTENCE>\newline
    Is the sentence supported by the context above?\newline
    Answer only in Yes or No:}
  \end{tabular} \\
\midrule
  \textbf{\texttt{Natural language evaluation using LLM}} &
  \begin{tabular}[c]{@{}p{5.2in}@{}}
  \textbf{\texttt{convince-factor-wth-answer\dag}}
  \\\\
  \texttt{Given the question: <QUESTION>\newline
    Given the answer: <ANSWER>\newline
    Given the reason: <REASON>\newline
    Please score how much the reason convinces you from 1 (not convinced) to 5 (very convinced):}
  \\\\
  \textbf{\texttt{convince-factor-without-answer\dag}}
  \\\\
  \texttt{Given the question: <QUESTION>\newline
    Given the reason: <REASON>\newline
    Please score how much the reason convinces you from 1 (not convinced) to 5 (very convinced):}
  \\\\
  \textbf{\texttt{factuality}\ddag}
  \\\\
  \texttt{Given the source document: <SOURCE>\newline
    Please score the factuality of the source document from 1 (not factually correct) to 5 (fully factually correct):}
   \\\\
   \textbf{\texttt{information-mismatch}\ddag}
   \\\\
   \texttt{Given the source document: <SOURCE>\newline
    Given the model-generated text: <GENERATED>\newline
    Please score the amount of information mismatch between source document and model-generated text from 1 (very less mismatch) to 5 (very high mismatch):}
    
  \end{tabular}\\
\bottomrule
\end{tabular}%
}
\end{center}
\caption{Prompts for 'SelfCheckGPT with Prompt' hallucination scoring scheme \& all modes under leveraging of GPT-3.5 Turbo \& GPT-4 Turbo for evaluation various attributes of main response reasoning passages. We use GPT-4 Turbo to verify responses from GPT-4 Turbo itself. Human evaluators are also provided with exactly same prompts.\\
\dag - \texttt{<QUESTION>}, \texttt{<ANSWER>} and \texttt{<REASON>} masks are replaced by the current question, main response answer and main response reasoning passage.\\
\ddag - \texttt{<SOURCE>} and \texttt{<GENERATED>} are replaced by main response reasoning passage and stochastic response reasoning passages.
}
\label{tab:prompting_templates_2}
\end{table*}

We shall now describe the exact prompts that we used.
\subsection{Collection of Closed Questions}
\label{sec: appendix-closed-questions}
We collect closed questions by prompting GPT-4 Turbo to create boolean problems from the passage given in the prompt. The passage is taken from the \texttt{wikipedia} pages of topics under different subjects.
Table \ref{tab:prompting_templates_1} shows the exact prompt that we used for collecting closed questions for \texttt{SCiPS-QA}. We replace the \texttt{<PASSAGE>} placeholder with the passages retrieved from \texttt{wikipedia}.

We observed that most of the questions created by GPT-4 Turbo in this manner, we purely a test of knowledge retrieval. This made us include some additional instructions in the prompt. We manually checked the questions for their corresponding answers and ensured that most of the questions in \texttt{SCiPS-QA} required some levels of reasoning to answer.
\subsection{Collecting Responses}
\label{sec: appendix-collect-response}
We now describe the prompts that we used for collecting responses from open-source models and proprietary models. 
\subsubsection{Open-source Models}
\label{sec: appendix-collect-response-open}
Table \ref{tab:prompting_templates_1} shows the exact prompts that we used for collecting responses (A- Yes, B - No \& C - I do not know) from open-source models. Table \ref{tab: perc_empty} shows the number of responses from each open-source model that were invalid. A response (main or stochastic) is considered to be invalid if it could not parsed into one of the choices (A, B or C). We observe that low-scale models Llama-2-7B, Llama-3-8B and Mistral-7B-Instruct-v0.1 had a high percentage of invalid main responses. The instruction fine-tuned versions of models reported much lesser invalid responses at same scale of parameters. The GPT line of models and higher scaled members of Llama-2 and Llama-3 family reported much less percentage of invalid responses (both 'main' and 'stochastic'). 

While collecting responses from open-source models, we set the generation parameter \texttt{max\_new\_tokens} to 3 and parse the responses for options from the set \{A, B, C\}, (A - "Yes", B - "No", C - "I do not know"). For models : Llama-2-70B, Llama-2-70B-chat, Llama-3-70B and Llama-3-70B-instruct, we use non-uniform 4-bit quantization to fit these models within a single A100 to account for limited computational resources. Since we also collect reasoning passages from chosen proprietary models, we set the generation parameter \texttt{max\_tokens} to 1000.
\subsubsection{Proprietary Models}
\label{sec: appendix-collect-response-proprietary}
The prompt structure for proprietary models differs from that for open-source models with respect to the presence of 'Reason:' field in the exemplars. This is done to force these models to provide reasoning passages which are further quantified for hallucination and score for different attributes using human experts and GPT-3.5 Turbo as evaluators in a parallel setting.

\subsection{SelfCheckGPT with Prompt}
\label{sec: appendix-selfcheckgpt-with-prompt}
Table \ref{tab:prompting_templates_2} shows the exact prompt that we used for this variant of SelfCheckGPT. The prompt is exactly same mentioned in the SelfCheckGPT paper \cite{manakul2023selfcheckgpt}.
The \texttt{<CONTEXT>} is replaced by each of the stochastic response passages and \texttt{<SENTENCE>} is replaced by the main reasoning passages.

\subsection{NLG Evaluation of Reasoning Passages}
\label{sec: appendix-nlg-eval-prompt}
We describe all the prompts that we used for this section. Note that we used GPT-4 Turbo, GPT-3.5 Turb and text-davinci-003 as the LLM modules for assigning scores to the main response reasoning passages.
\label{sec: appendix-nlg-eval}
\subsubsection{Convince-factor}
Table \ref{tab:prompting_templates_2} shows the prompt that we used for two schemes : \texttt{convince-factor-with-answer} and \texttt{convince-factor-without-answer}. The two prompts differed only with respect to the presence of the model answer (to the boolean scientific query). 
\subsubsection{Fact-check}
Table \ref{tab:prompting_templates_2} shows the prompt that we used for assessing the factuality of main response reasoning passages (which replaced the \texttt{<SOURCE>} placeholder)
\subsubsection{Information-mismatch}
Table \ref{tab:prompting_templates_2} shows the prompt that we used for assigning scores of this attribute. The \texttt{<SOURCE>} placeholder is replaced with the main response reasoning passage and the <GENERATED> placeholder is replaced with the stochastic response reasoning passages. 

\section{Results}
\subsection{Invalid Responses}
\label{sec:appendix-perc-empty}
Table \ref{tab: perc_empty} shows the percentage of invalid responses (to the '\texttt{answer}' field of the prompt) to queries in \texttt{SCiPS-QA}. Llama-3 models and GPT models show fairly low numbers of invalid responses.
Low scale models from Llama-2, Llama-3 and Mistral family report high percentage of invalid responses.
\label{sec: appendix-result-benchmark}
\begin{table*}[h]
\begin{center}
\resizebox{\textwidth}{!}{%
\begin{tabular}{l|c|c}
\textbf{models}            & \textbf{Percentage invalid main responses} & \textbf{Average percentage invalid stochastic responses} \\
\toprule
meta-llama-2-7B            & 1.000                                          & 0.558                                                    \\
meta-llama-2-7B-chat       & 0.152                                      & 0.271                                                    \\
meta-llama-2-13B           & 0.008                                      & 0.239                                                    \\
meta-llama-2-13B-chat      & 0.026                                      & 0.099                                                    \\
meta-llama-2-70B           & 0.136                                      & 0.154                                                    \\
meta-llama-2-70B-chat      & 0.019                                      & 0.4                                                      \\
meta-llama-3-8B            & 0.753                                      & 0.45                                                     \\
meta-llama-3-8B-instruct   & 0.001                                      & 0.046                                                    \\
meta-llama-3-70B           & 0.005                                      & 0.219                                                    \\
meta-llama-3-70B-instruct  & 0.008                                      & 0.029                                                    \\
Mistral-7B-Instruct-v0.1   & 0.693                                      & 0.339                                                    \\
Mistral-7B-Instruct-v0.2   & 0.089                                      & 0.138                                                    \\
Mixtral-8x7B-Instruct-v0.1 & 0.034                                      & 0.113                                                    \\
text-davinci-003           & 0.011                                      & 0.011                                                    \\
GPT-3.5 Turbo              & 0.005                                      & 0.011                                                    \\
GPT-4 Turbo                & 0.000                                          & 0.002 \\
\bottomrule
\end{tabular}
}
\caption{Percentage invalid responses across all open-source \& proprietary models. Low scale models : meta-llama-2-7B \& Mistral-7B-Instruct-v0.1 report highest percentage of invalid main responses. GPT models report lowest percentage invalid responses.}
\label{tab: perc_empty} 
\end{center}
\end{table*}

\subsection{Hallucination Quantification}
\label{sec: appendix-result-hall-quant}
\begin{figure}[ht!]
\begin{center}
\begin{tabular}{c}
    \includegraphics[width = \linewidth]{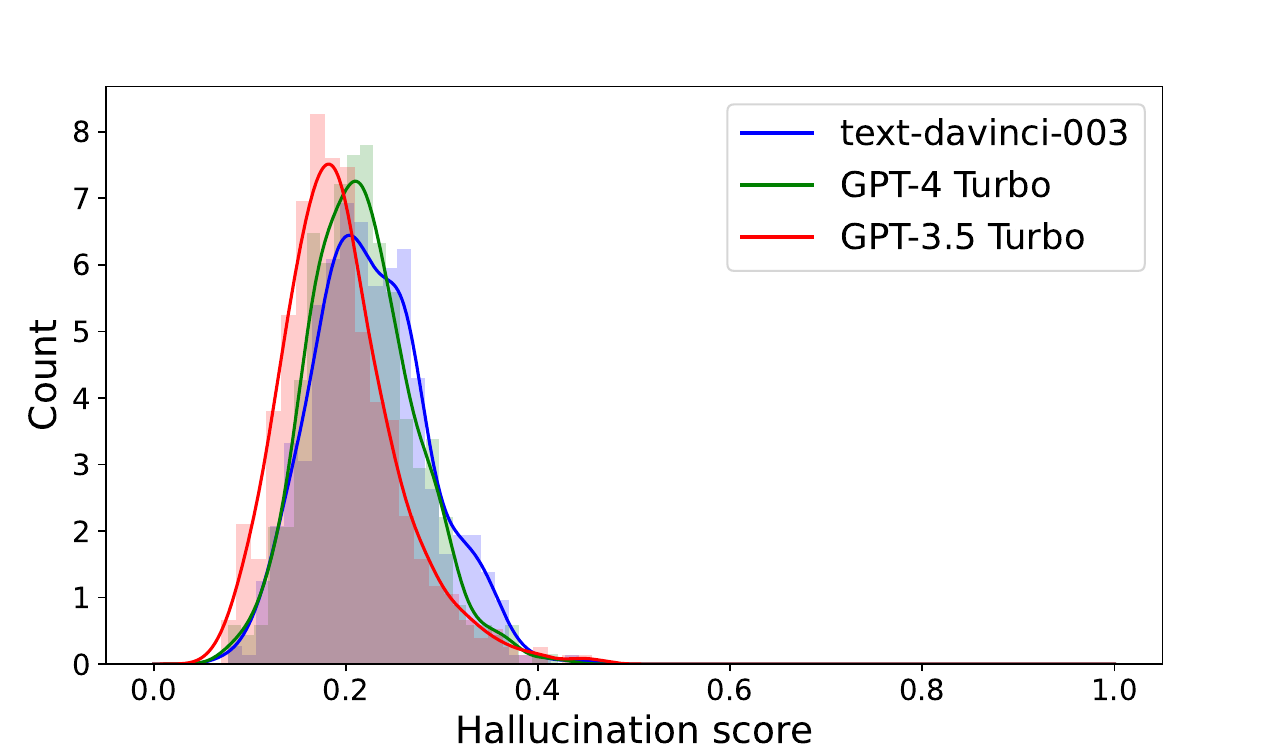} 
    \\
    \includegraphics[width = \linewidth]{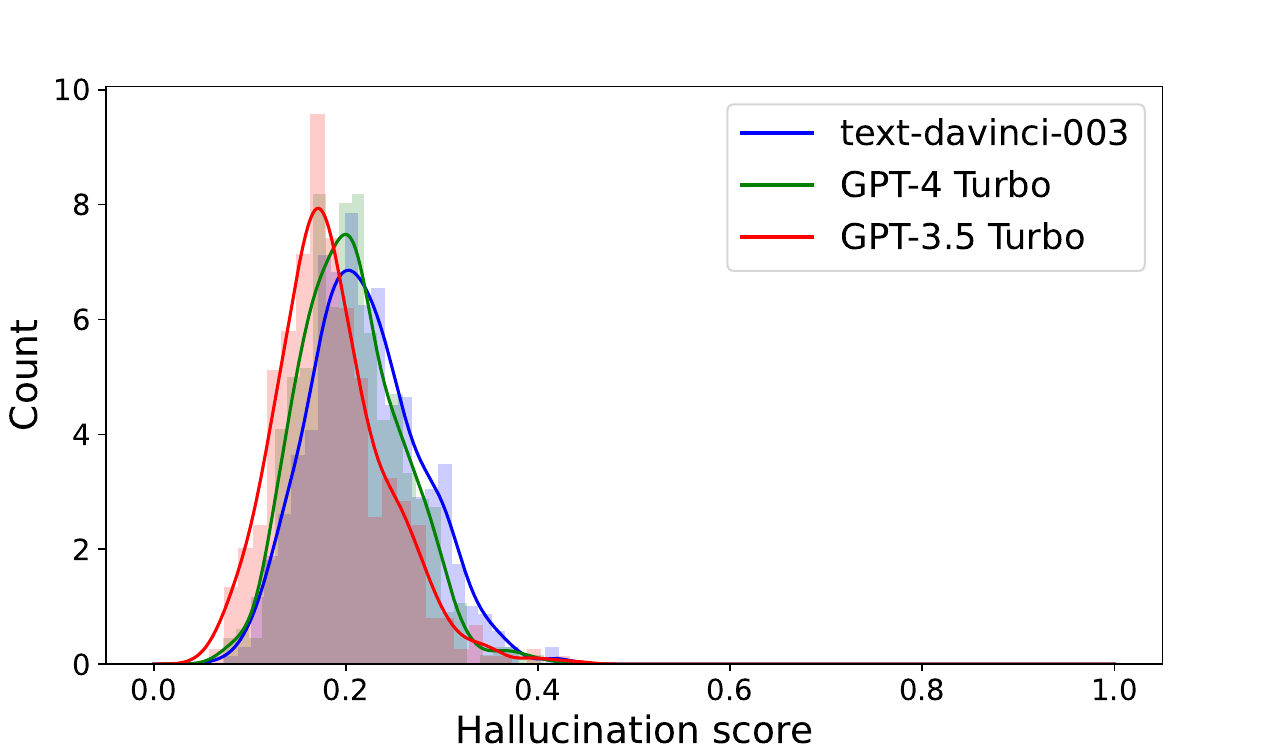}
\end{tabular}

\end{center}
\caption{Frequency distribution plots of `SelfCheckGPT with BERTScore' hallucination scores to main response reasoning passages for sentence\_transformer models: all-MiniLM-L6-v2 (above) \& all-mpnet-base-v2 (below)}
\label{fig:result_bert_score}
\end{figure}

\begin{figure}[t!]
    \centering
    \includegraphics[width = \linewidth]{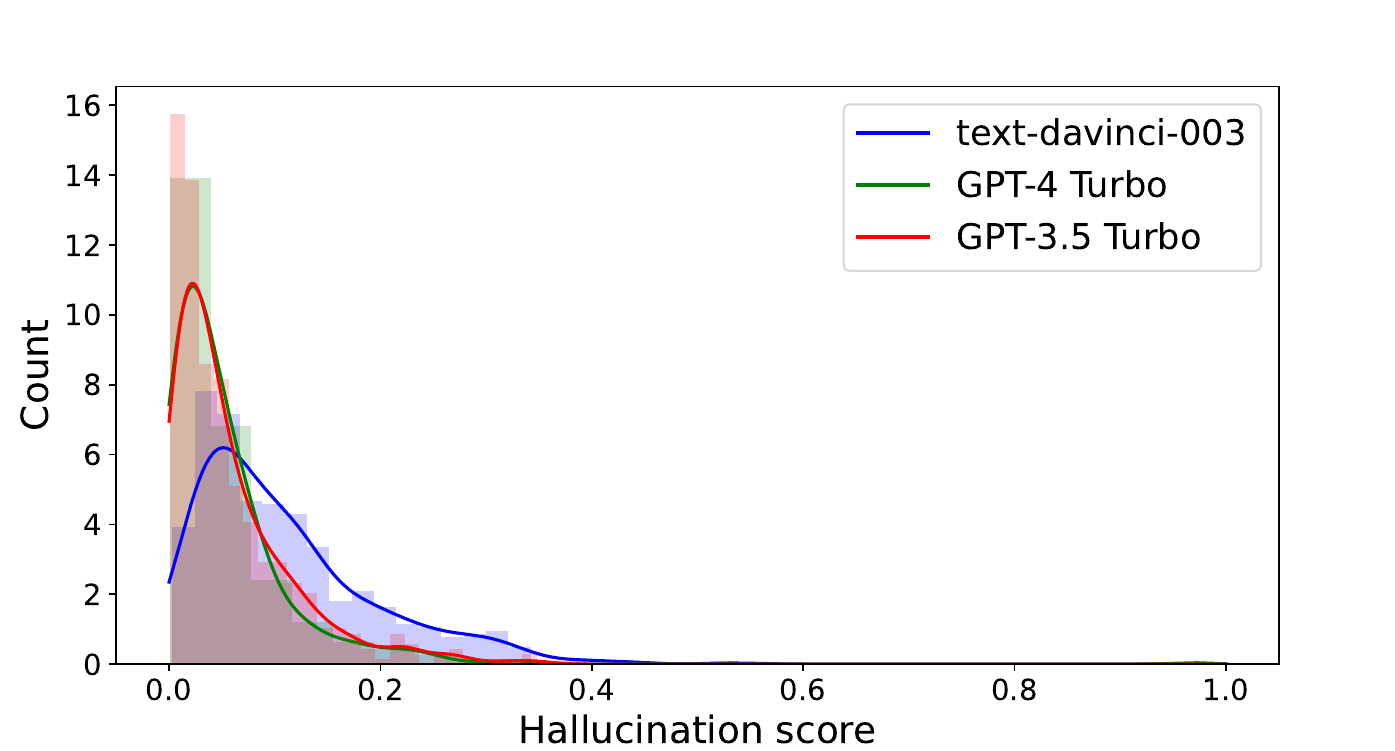}
    \caption{Frequency distribution plots of 'SelfCheckGPT with NLI' hallucination scores to the main response reasoning passages.}
    \label{fig:selfcheckgpt_nli_result}
\end{figure}

\begin{figure}[t!]
    \centering
    \includegraphics[width = \linewidth]{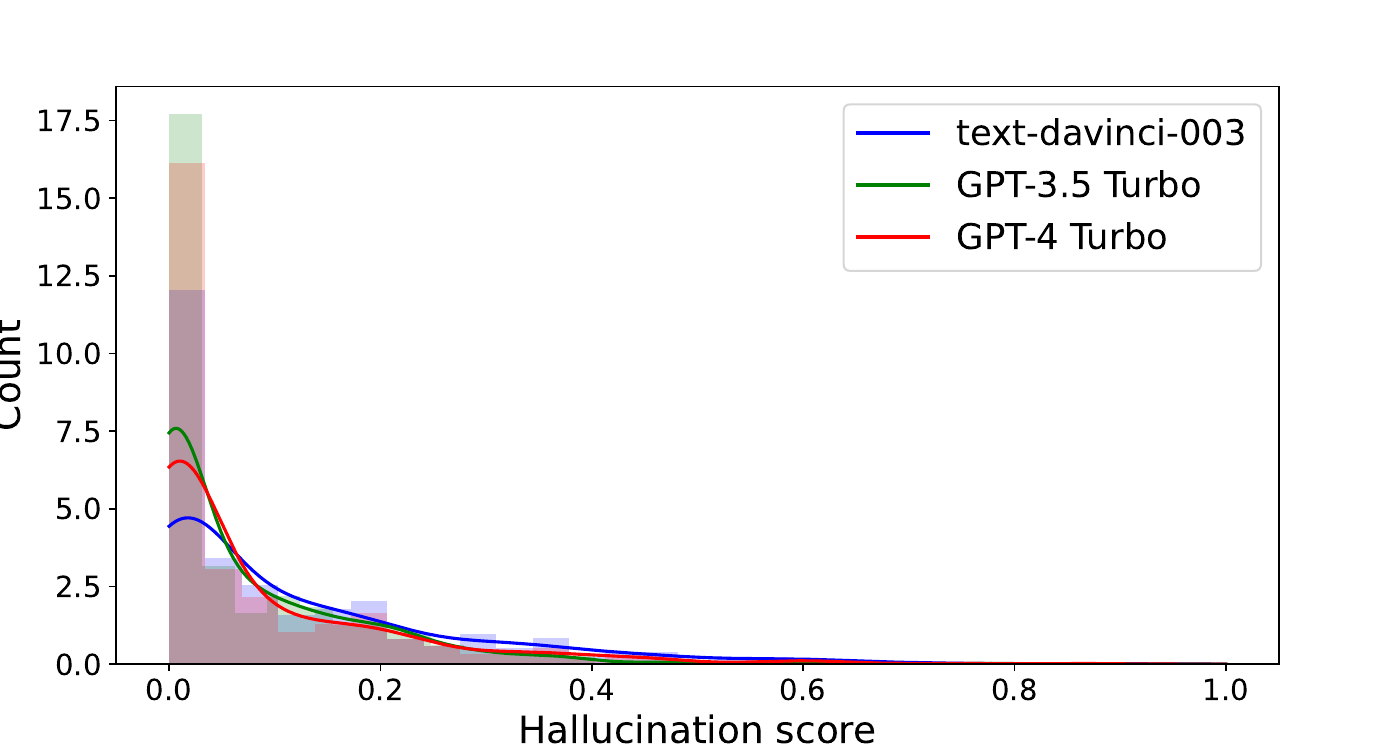}
    \caption{Frequency distribution plots of 'SelfCheckGPT with Prompt' hallucination scores to the main response reasoning passages.}
    \label{fig:selfcheckgpt_prompt_result}
\end{figure}

\subsubsection{SelfCheckGPT with NLI} Figure \ref{fig:selfcheckgpt_nli_result} shows that main response passages from GPT-3.5 Turbo and GPT-4 Turbo are not demarcated for amount of hallucination using this scoring. text-davinci-003 however, is clearly shown to produce more hallucinated text.

\subsubsection{SelfCheckGPT with Prompt} Figure \ref{fig:selfcheckgpt_prompt_result} shows results for SelfCheckGPT with Prompt. More response passages from GPT-3.5 Turbo are given low hallucination scores as compared to those from GPT-4 Turbo.

\subsubsection{SelfCheckGPT with BERTScore}
We performed Welch's t-tests to test the statistical significance of results. We observe that main response reasoning passages from GPT-3.5 Turbo are given least mean hallucination scores using 'SelfCheckGPT with BERTScore' and main response reasoning passages from text-davinci-003 are given the highest mean hallucination scores. We confirm this with Welch's t-tests conducted using \texttt{scipy.stats.ttest\_ind} : 
\\
\textbf{Notation} : Let $\mu_{gpt-4-turbo}$, $\mu_{gpt-3.5-turbo}$ \& $\mu_{text-davinci-003}$ represent the sample means of hallucination scores. The details of the tests are present in Table \ref{tab:hypothesis}
\begin{table*}[t]
\centering
\resizebox{\textwidth}{!}{%
\begin{tabular}{cccccc}

\textbf{Models} & \textbf{$H_{0}$ (Null Hypothesis)} &\textbf{ $H_{1}$ (Alternate Hypothesis)} & \textbf{p-value} & \textbf{degrees of freedom (df)} & \textbf{Result} \\ \hline\\

\multirow{3}{*}{' ' \textbf{all-MiniLM-L6-v2}} & 
 $\mu_{\text{gpt-4-turbo}} = \mu_{\text{gpt-3.5-turbo}}$ & 
 $\mu_{\text{gpt-4-turbo}} > \mu_{\text{gpt-3.5-turbo}}$ & 
 5.62e-10 & 
 979.96 & 
 Reject Null Hypothesis \\\\

 &  $\mu_{\text{gpt-4-turbo}} = \mu_{\text{text-davinci-003}}$ & 
 $\mu_{\text{gpt-4-turbo}} < \mu_{\text{text-davinci-003}}$ & 
 0.0037 & 
 979.87 & 
 Reject Null Hypothesis \\\\

 &  $\mu_{\text{text-davinci-003}} = \mu_{\text{gpt-3.5-turbo}}$ & 
 $\mu_{\text{text-davinci-003}} > \mu_{\text{gpt-3.5-turbo}}$ & 
 3.55e-17 & 
 985.99 & 
 Reject Null Hypothesis \\\\  
 
 \hline\\
 
 \multirow{3}{*}{' ' \textbf{all-mpnet-base-v2}} & 
 $\mu_{\text{gpt-4-turbo}} = \mu_{\text{gpt-3.5-turbo}}$ & 
 $\mu_{\text{gpt-4-turbo}} > \mu_{\text{gpt-3.5-turbo}}$ & 
 4.08e-09 & 
 981.86 & 
 Reject Null Hypothesis \\\\

&  $\mu_{\text{gpt-4-turbo}} = \mu_{\text{text-davinci-003}}$ & 
 $\mu_{\text{gpt-4-turbo}} < \mu_{\text{text-davinci-003}}$ & 
 2.58e-05 & 
 980.16 & 
 Reject Null Hypothesis \\\\

&  $\mu_{\text{text-davinci-003}} = \mu_{\text{gpt-3.5-turbo}}$ & 
 $\mu_{\text{text-davinci-003}} > \mu_{\text{gpt-3.5-turbo}}$ & 
 5.28e-21 & 
 985.85 & 
 Reject Null Hypothesis \\\\
 \hline
  \end{tabular}
}
\caption{Welch's t-tests for testing difference in means of hallucination scores given to main response reasoning passages under SelfCheckGPT with BERTScore method. The level of significance for all these tests is 0.05.\\Note: We assumed the normality of distribution of the hallucination scores for each of the proprietary model and we did not assume anything about their variances.}
\label{tab:hypothesis}
\end{table*}

\begin{table*}[ht!]
\begin{center}
\resizebox{1\textwidth}{!}{%
\begin{tabular}{m{10cm}m{25cm}m{25cm}m{25cm}m{1cm}}
    \toprule
    \\
    \\
    \centering \fontsize{40pt}{40pt}\selectfont \textbf{attribute} &\centering \fontsize{40pt}{40pt}\selectfont \textbf{GPT-4 Turbo} & \centering \fontsize{40pt}{40pt}\selectfont \textbf{GPT-3.5 Turbo} & \centering \fontsize{40pt}{40pt}\selectfont \textbf{text-davinci-003}& \\\\\\
    \toprule
    \\
    \\
    \centering \textbf{\fontsize{40pt}{40pt}\selectfont convince-factor-with-answer }
    & 
    \includegraphics[width = \linewidth]{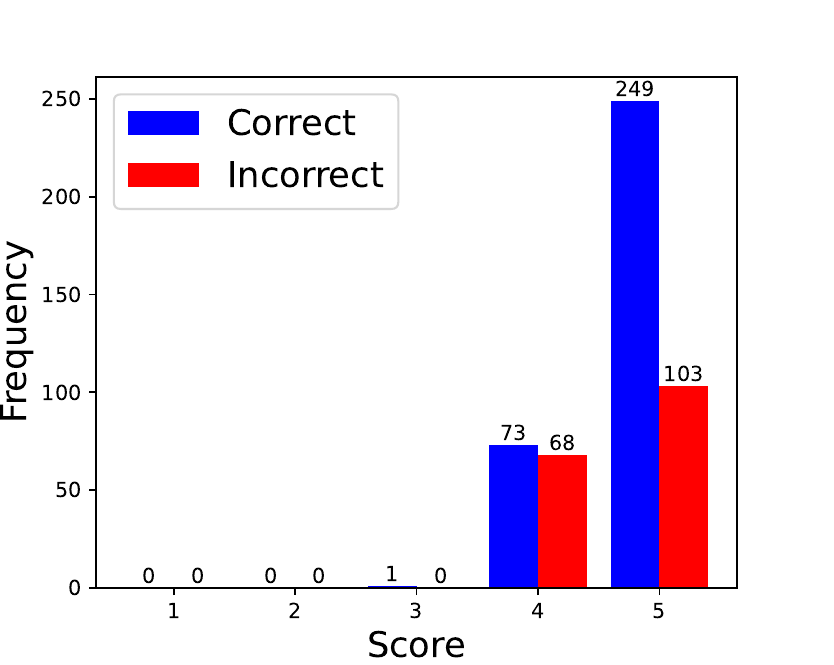} 
    &
    \includegraphics[width = \linewidth]{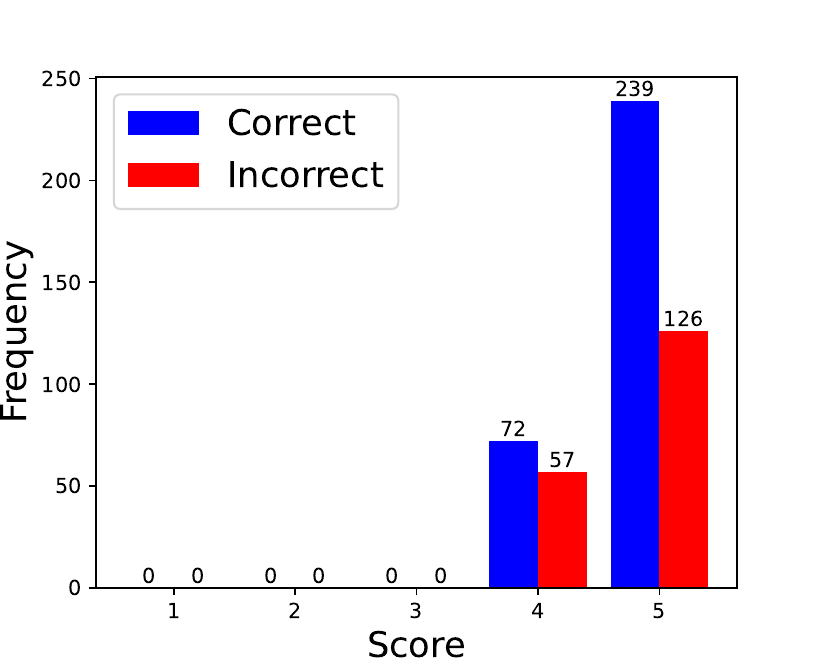} 
    &
    \includegraphics[width = \linewidth]{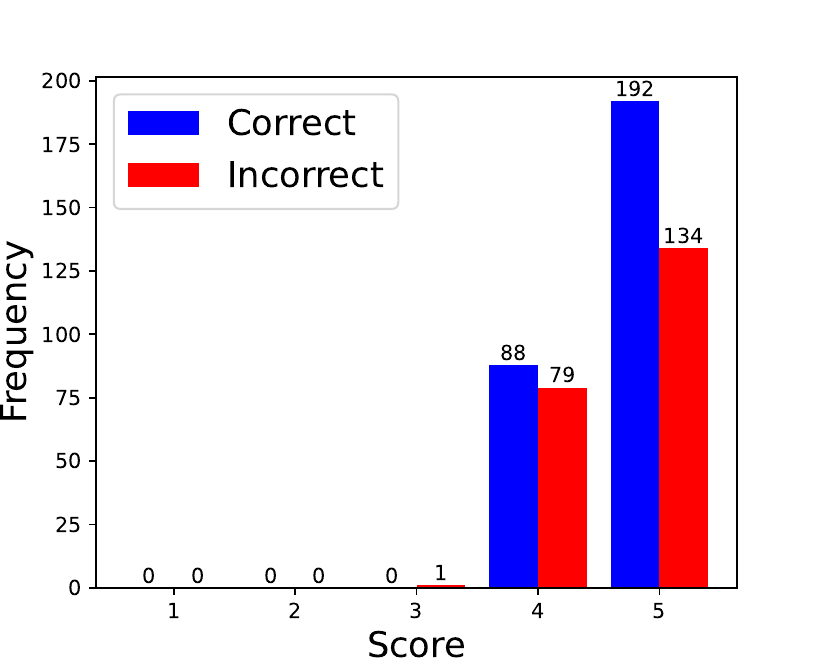}\\
    
    \centering \textbf{\fontsize{40pt}{40pt}\selectfont convince-factor-without-answer }
    & 
    \includegraphics[width = \linewidth]{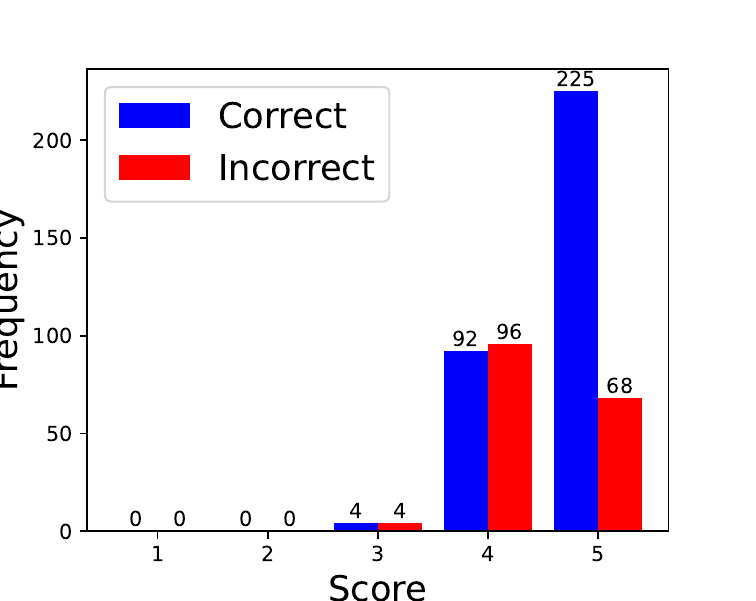} 
    &
    \includegraphics[width = \linewidth]{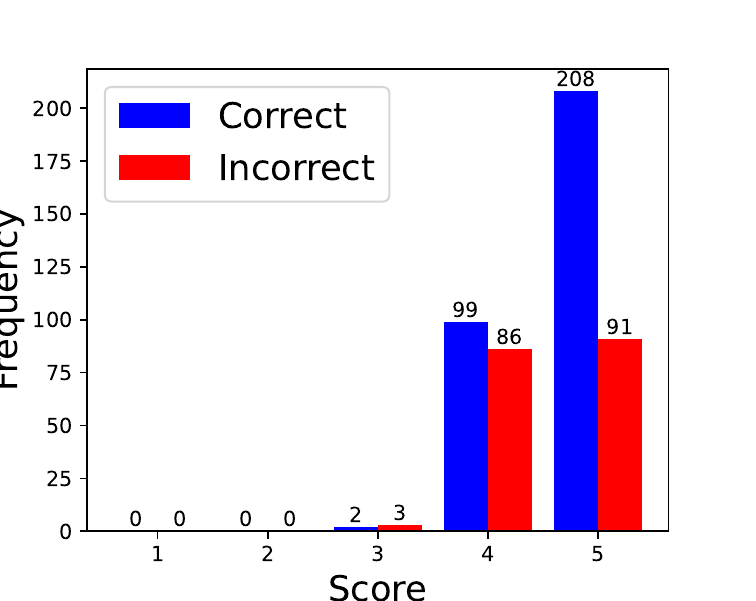} 
    &
    \includegraphics[width = \linewidth]{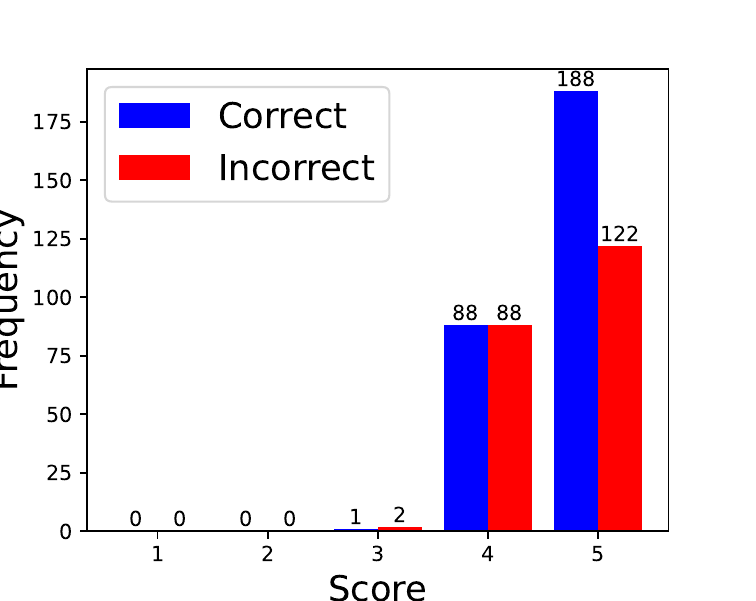}\\
    
    \centering \textbf{\fontsize{40pt}{40pt}\selectfont fact-check}
    & 
    \includegraphics[width = \linewidth]{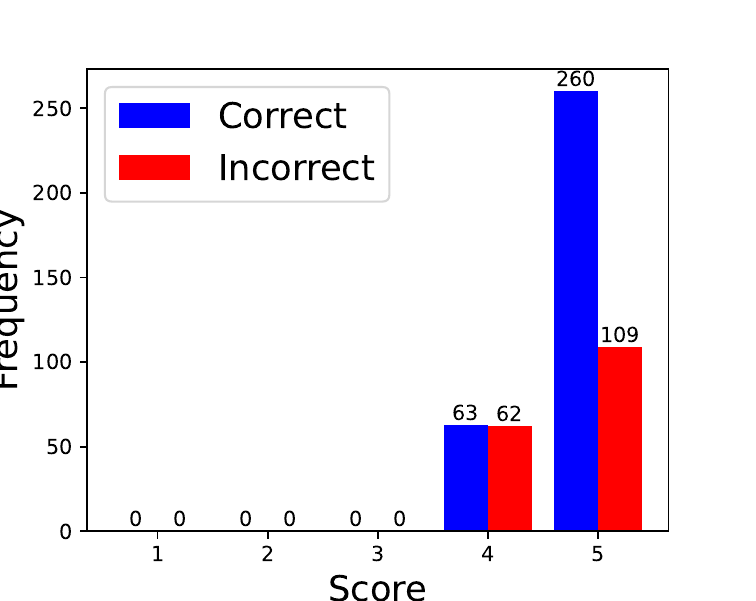} 
    &
    \includegraphics[width = \linewidth]{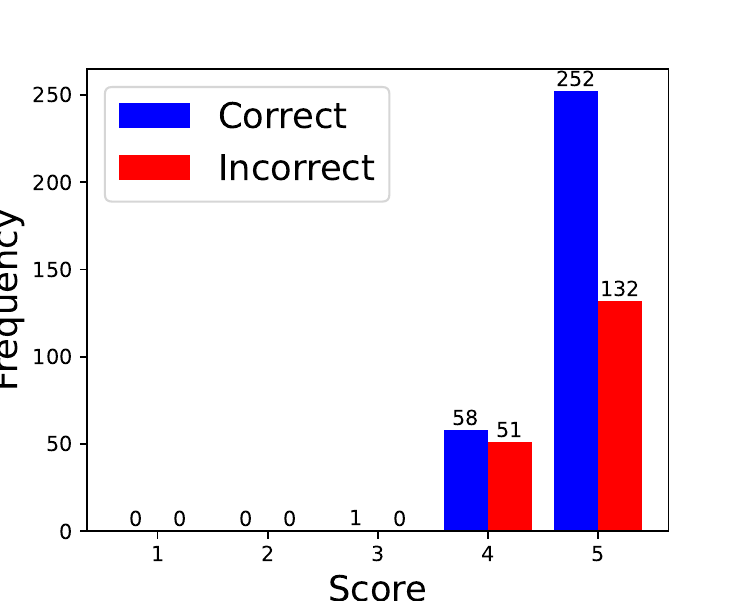} 
    &
    \includegraphics[width = \linewidth]{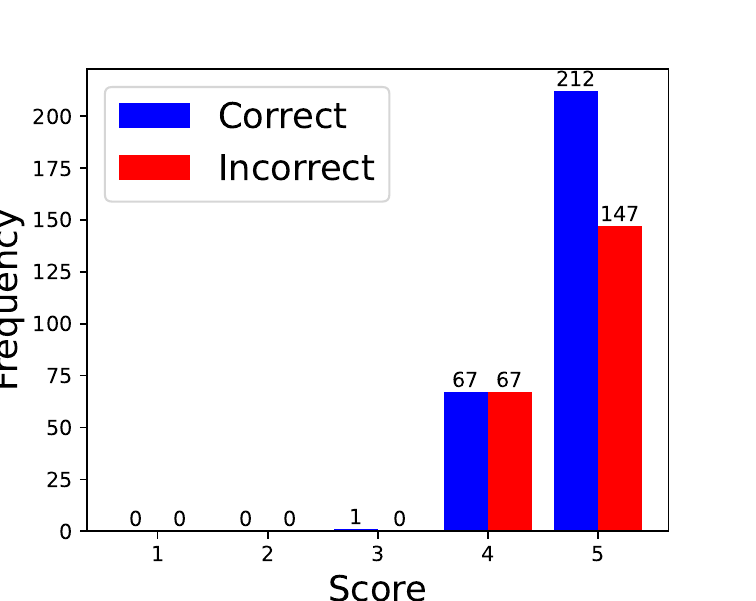}\\
    
    \centering \textbf{\fontsize{40pt}{40pt}\selectfont information-mismatch} 
    & 
    \includegraphics[width = \linewidth]{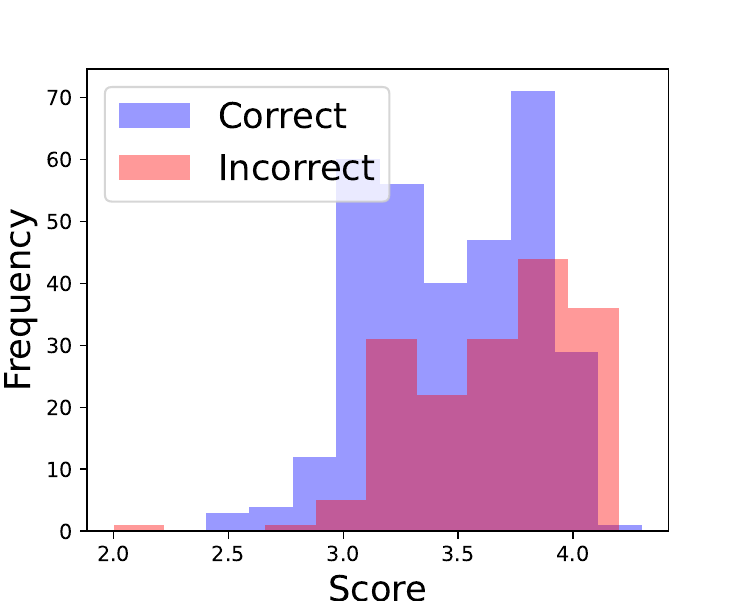} 
    &
    \includegraphics[width = \linewidth]{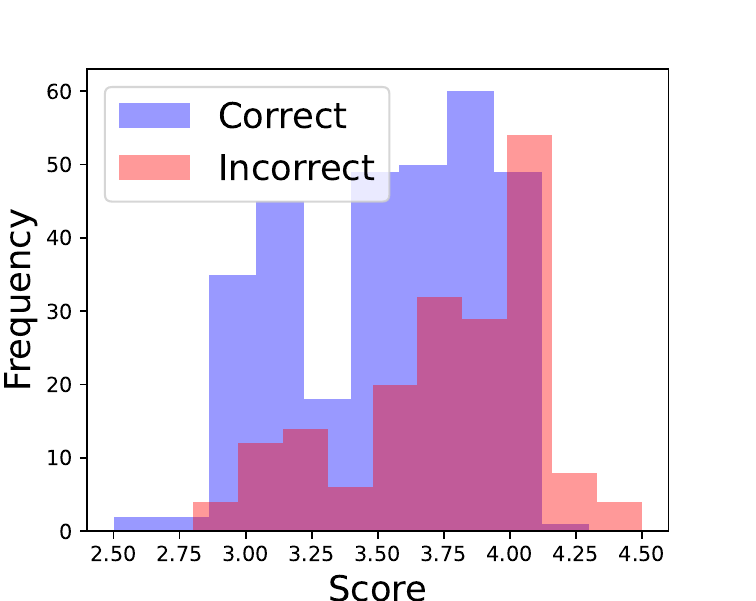} 
    &
    \includegraphics[width = \linewidth]{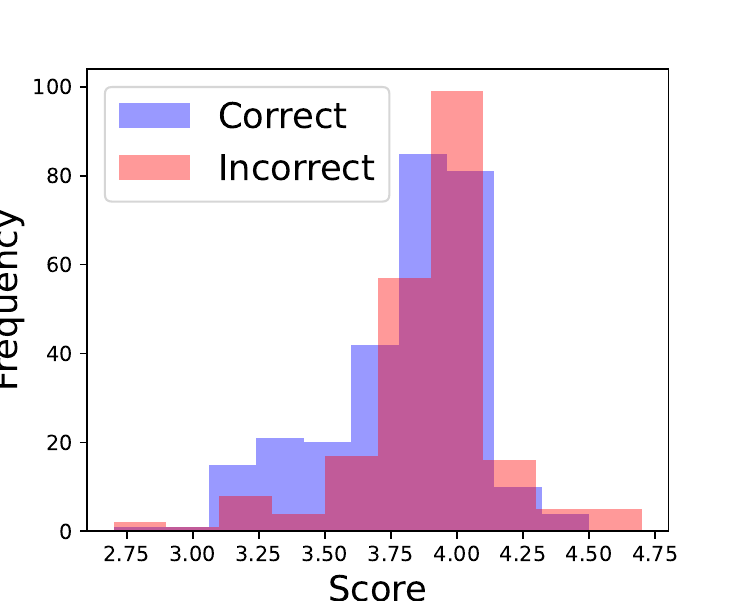}
\end{tabular}
}
\end{center}
\caption{Verification of the main response reasoning passages generated by all three proprietary models across convincingness (with and without answer), factuality, and information mismatch using GPT-3.5 Turbo as the verifier model.}
\label{tab:nlg_evaluation_result}
\end{table*}

\end{document}